\documentclass[11pt]{article}

\usepackage{PRIMEarxiv}
\RequirePackage{algorithm}
\RequirePackage{algorithmic}

\usepackage{amsmath}
\usepackage{amssymb}
\usepackage{xcolor} 
\usepackage{makecell}
\usepackage{subcaption}
\usepackage[utf8]{inputenc} 
\usepackage[T1]{fontenc}    
\usepackage{hyperref}       
\usepackage{url}            
\usepackage{booktabs}       
\usepackage{amsfonts}       
\usepackage{nicefrac}       
\usepackage{microtype}      
\usepackage{lipsum}
\usepackage{fancyhdr}       
\usepackage{graphicx}       
\graphicspath{{..}}

\title{Finite Expression Methods for Discovering Physical Laws from Data 
}

\author{
  Zhongyi Jiang \\
  Department of Statistics and Data Science\\
  Yale University \\
  New Haven, CT, USA\\
  \texttt{zhongyi.jiang@yale.edu} \\ 
   \And
  Chunmei Wang \\
  Department of Mathematics \\
  University of Florida \\
  Gainesville, FL, USA\\
  \texttt{chunmei.wang@ufl.edu} \\
    \And
  Haizhao Yang\thanks{Corresponding author.}\\
  Department of Mathematics \\
  Department of Computer Science\\
  University of Maryland College Park\\
  College Park, MD, USA\\
  \texttt{hzyang@umd.edu}
}

\begin{document}
\maketitle

\begin{abstract}Nonlinear dynamics is a pervasive phenomenon observed in scientific and engineering disciplines. However, the task of deriving analytical expressions to model nonlinear dynamics from data remains challenging. In this paper, we shall present a novel deep symbolic learning method called the "finite expression method" (FEX) to discover governing equations within a function space containing a finite set of analytic expressions, based on observed dynamic data. The key concept is to employ FEX to generate analytic expressions of the governing equations by learning the derivatives of partial differential equation (PDE) solutions through convolutions. Our numerical results demonstrate that our FEX surpasses other existing methods (such as PDE-Net, SINDy, GP, and SPL) in terms of numerical performance across a range of problems, including time-dependent PDE problems and nonlinear dynamical systems with time-varying coefficients. Moreover, the results highlight FEX's flexibility and expressive power in accurately approximating symbolic governing equations.
\end{abstract}




\section{Introduction}
Partial differential equations and dynamical systems are fundamental tools in describing the laws of physics in different scientific and engineering domains. However, comprehending the underlying mechanisms of complex systems in modern applications, such as climate science, neuroscience, ecology, finance, and epidemiology, remains a challenge. Traditionally, empirical forms have been used to derive the governing equations for these systems ~\cite{efendiev2018mathematical,brennen2005fundamentals}. In recent years, the rapid advancement of computational power and the availability of vast amounts of observational and stored data have opened up new opportunities for data-driven methodologies to establish physical laws \cite{SChen, GONZALEZGARCIA1998S965,Raissi2017PhysicsID,Lagergren2020BiologicallyinformedNN,Lusch,doi:10.1098/rspa.2017.0844,QIN2019620,HARLIM2021109922,bongard2007automated, schmidt2009distilling, brunton2016discovering, wsindy, rudy2017data, schaeffer2017learning, Sindy2, wu2019learning, schaeffer2020extracting, chang2019identification,doi:10.1137/20M134513X,ident,MESSENGER2021110525}. These data-driven approaches offer an alternative approach to discover governing equations by harnessing the wealth of information present in the data. By combining advanced computational techniques with data observation and analysis, researchers can explore and understand the intricate behavior of these systems more comprehensively and insightfully.

In the field of discovering physical laws through data-driven methods, researchers have made significant contributions since the early works \cite{SChen, GONZALEZGARCIA1998S965}. In recent years, the significant progress in machine learning, data science, and computing power has enabled groundbreaking efforts to uncover the governing equations of physical systems. When the underlying physical system is known in advance, deep learning methods have been developed to enhance modeling accuracy by learning data-adaptive system parameters alongside the known equations \cite{Raissi2017PhysicsID, Lagergren2020BiologicallyinformedNN}. However, the more challenging task arises when attempting to identify the complete physical system from data, without any prior knowledge. One approach is to utilize deep learning as a black-box regression tool for data-driven modeling and prediction \cite{Lusch,doi:10.1098/rspa.2017.0844,QIN2019620,HARLIM2021109922}. Although this approach allows for theoretical analysis and accuracy guarantees \cite{Tu2020Understanding, doi:10.1137/21M140691X}, it falls short in terms of analytically identifying the governing equations and, consequently, lacks interpretability. 

Symbolic regression has emerged as a promising methodology in recent years for analytically identifying governing equations from data, offering improved interpretability  \cite{wuxiu1, wuxiu2, schmidt2009distilling, brunton2016discovering, rudy2017data, schaeffer2017learning, Sindy2, wu2019learning, schaeffer2020extracting, chang2019identification, Su, Xu, Fu, Chen, Churchill, Churchill2}.   
One popular symbolic approach involves constructing a comprehensive dictionary of mathematical symbols and learning a sparse linear combination of these symbols to approximate the desired governing equation through a least squares problem. For example, the Sparse Identification of Nonlinear Dynamics (SINDy)  introduced by Brunton et al. \cite{brunton2016discovering}  has provided significant insights into addressing this challenge. SINDy employs a sequential threshold ridge regression algorithm to recursively determine a sparse solution from a predefined library of basis functions. However, the success of this sparsity-promoting approach heavily relies on a well-defined candidate function library, which requires prior knowledge of the system. Furthermore, a linear combination of candidate functions may not be sufficient to capture complex mathematical expressions. When dealing with a massive library size, the sparsity constraint is empirically observed to fail, and the resulting high memory and computational costs become prohibitive, particularly in high-order and high-dimensional systems.

Another approach in symbolic regression involves the design of symbolic neural networks to generate complex mathematical expressions through the linear combination and multiplication of mathematical symbols. One method in this category is PDE-Net
\cite{long2019pde,long2018pde}. The effectiveness of symbolic neural networks in generating desired governing equations is determined by the depth and width of the network. Network depth plays a crucial role in capturing the nonlinearity of governing equations and needs to be sufficiently large to uncover highly nonlinear relationships. However, optimizing symbolic neural networks poses a challenging problem, leading to reduced effectiveness of the approach. 
Furthermore, the symbolic neural networks utilized in PDE-Net are limited to representing polynomials of mathematical operators and do not consider time-varying coefficients. This restriction confines their applicability to a specific class of physical laws, which limits their use where time-varying coefficients are essential.

The symbolic regression methods such as genetic algorithms (GA) \cite{forrest1993genetic} and genetic programming (GP) \cite{koza1994genetic}, have been recently explored for discovering governing equations. GP, inspired by Darwin's theory of evolution, has been proposed to identify approximate governing equations \cite{wang2019symbolic} through evolving mathematical expressions to better fit given data. GP represents mathematical expressions as tree-structured chromosomes consisting of nodes and terminals, while GA represents chromosomes as strings of binary digits.
During the training process, GP employs random crossover and mutation operations on mathematical expressions to evolve generations of expressions. However, due to the discrete nature of the optimization process in GP, there is a lack of continuous optimization for fine-tuning parameters. As a result, GP faces difficulties in providing highly accurate mathematical expressions, even for cases involving constant coefficients. 

Reinforcement learning (RL) was proposed to identify mathematical expressions, including governing equations \cite{petersen2021deep, sun2023symbolic, liang2022finite}.  However, the existing literature primarily focuses on numerical tests involving polynomial functions or simple dynamical systems with constant coefficients.  A recent method called Symbolic Physics Learner (SPL) \cite{sun2023symbolic} has introduced the use of Monte Carlo tree search (MCTS)  for discovering governing equations. SPL formulates the problem of symbolic regression within the framework of MCTS, where mathematical expressions containing operators and operands are generated concurrently by an agent. Only expressions yielding higher rewards are retained to update the agent. However, due to the simultaneous selection of operators and operands, SPL requires a large pool of candidate operators and operands. During its optimization process, SPL selects operands separately, resulting in the need for a large tree structure to represent the true expression. Consequently, solving the expression optimization problem becomes challenging.

In this paper, we explore an extension to the recently developed deep symbolic method called the finite expression method (FEX) introduced by Liang et al.  \cite{liang2022finite}, aiming to identify governing equations within the function space of mathematical expressions generated by binary expression trees with a fixed number of operators. The original FEX method was primarily designed to determine solutions for high-dimensional PDEs, where no learning mechanism exists to select derivatives to form governing equations. However, naively applying the original FEX necessitates a large number of derivatives for multiple variables within an operator list, resulting in significant challenges when solving the expression optimization problem, similar to the issues encountered in SPL.
To address this challenge, we adopt a data-driven approach that leverages continuous optimization to learn the appropriate derivatives within a governing equation. This process can be viewed as optimizing convolution kernels in convolutional neural networks  \cite{long2019pde}, where identifying suitable derivatives is analogous to optimizing these kernels. By employing continuous optimization techniques, we successfully reduce the complexity of learning a governing equation to that of learning a multivariate function. Consequently, FEX becomes an efficient solution for this purpose. 
For instance, the task of learning $\frac{du}{dt}+u\frac{du}{dx}=0$ is equivalent to learning a polynomial $a(u)+b(u)u$ involving functions $a(u)$, $b(u)$, and $u$ as variables, where $a$ and $b$ represent convolution operators that are used to implement derivatives (i.e., matrix-vector multiplication).

In summary, our FEX method introduces a compact tree structure for operands and utilizes unary operators which take a linear combination of distinct operands with trainable convolution parameters as input. This design effectively reduces the complexity of the optimization problem, offering a more manageable solution compared to SPL. Our learning agent focuses on the selection of operator sequences, while the selection of operands is learned through data-driven continuous optimization. As a result, FEX significantly simplifies the expression optimization problem and broadens the applicability of the RL-based symbolic regression method to the discovery of PDEs with varying coefficients.

\begin{figure*}[htbp]
    \centerline{\includegraphics[width=6in]{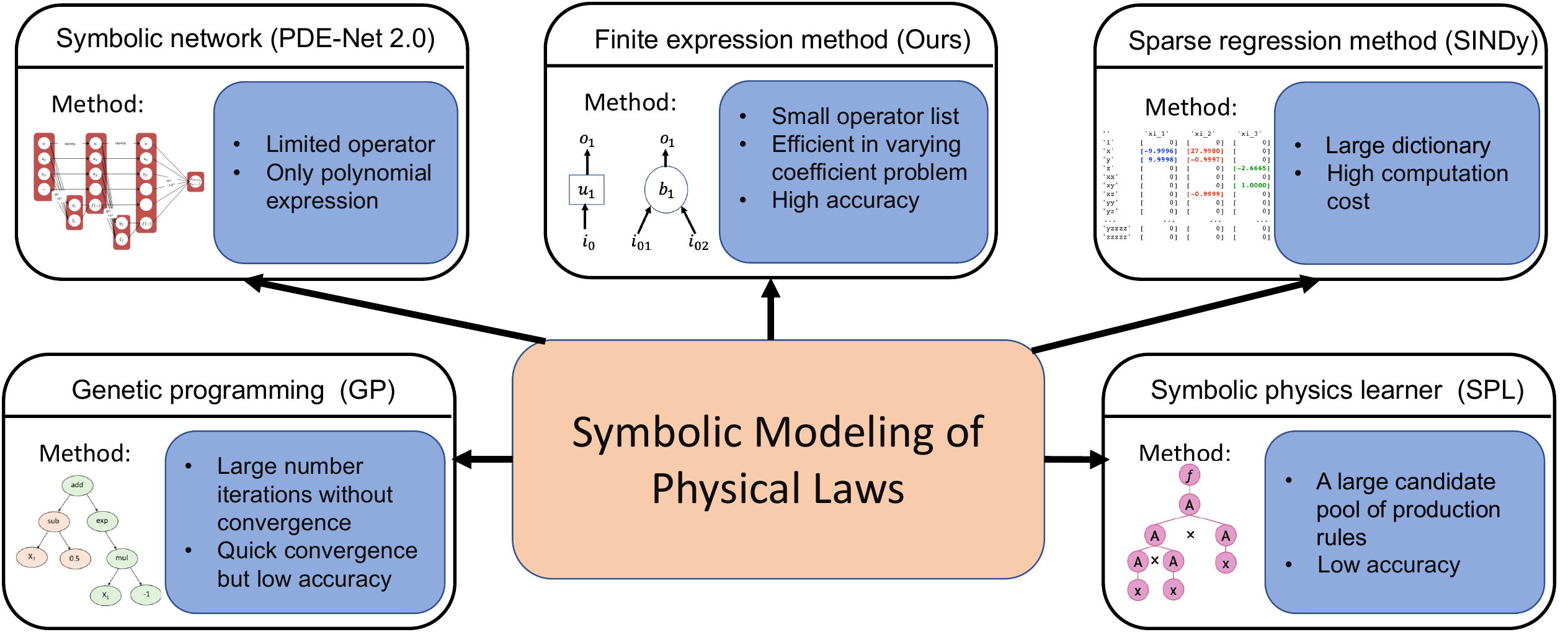}}
    \begin{center}
    \caption{Overview of different symbolic methods for discovering governing equations from data.}
    \label{FEX summary}
    \end{center}
\end{figure*}

Compared to other symbolic approaches for discovering physical laws, our FEX method offers several distinct advantages, as summarized in Figure ~\ref{FEX summary}: 
(1) \textbf{Efficient Expression Generation}: FEX utilizes binary expression trees to generate a wide range of mathematical expressions using a small operator list. This eliminates the need for a large dictionary of symbols or a complex symbolic neural network, as required by other methods  \cite{brunton2016discovering,Sindy2, long2019pde,long2018pde}. (2) \textbf{Improved Accuracy for Simple Equations}: FEX outperforms existing methods  \cite{brunton2016discovering,Sindy2,long2019pde,long2018pde,wang2019symbolic}, in terms of accuracy when learning simple equations with constant coefficients. (3) \textbf{Discovery of Highly Nonlinear Equations}: FEX excels in discovering highly nonlinear governing equations with varying coefficients where conventional symbolic approaches struggle to perform effectively. (4)  \textbf{Reformulation as Reinforcement Learning}: FEX formulates the problem of finding an appropriate mathematical expression as an RL  problem. This reformulation allows for the application of efficient optimization algorithms, resulting in high efficiency in terms of memory cost. Overall, our FEX exhibits promising potential for discovering complex systems and offers significant advantages over alternative symbolic approaches.

\section{FEX Methodology}\label{method}
\setlength{\footskip}{3.30003pt}
In this section, we shall first review the convolution filter method \cite{long2019pde} for computing derivatives in Section 2.1. We will present FEX method for discovering physical laws in Section 2.2.

\subsection{Schemes for Computing Derivatives}\label{derivatives computing} 
We will review the convolution filter method proposed by Long et al. \cite{long2019pde} for computing derivatives. The relation between convolutions and differentiations has been extensively studied \cite{cai2012image, dong2017image}. The study explores the connection between the order of sum rules of filters and the order of differential operators.
More precisely, the convention of the convolution operation on an image $f$ as a two-dimensional function aligns with the conventions commonly used in deep learning. The convolution operation can be defined as follows:
$$
(f \otimes q)\left[l_1, l_2\right]:=\sum_{k_1, k_2} q\left[k_1, k_2\right] f\left[l_1+k_1, l_2+k_2\right],
$$
where $\otimes$ represents circular convolution, $\left[\cdot, \cdot\right]$ specifies the pixel location of an image, and $q$ denotes the convolution filter as a matrix. It is important to note that this operation is a form of correlation rather than a strict mathematical convolution. For example, we set up convolution filters using a second-order upwind finite difference scheme. When the filter size is $5 \times 5$, the kernel matrices of the convolution filters for the corresponding discretizations 
$\frac{\partial }{\partial x}$ and $\frac{\partial^2 }{\partial x^2}$ for a two-dimensional function $u(x,y)$ are as follows:
$$
q_{01}=\begin{pmatrix} 0 &0 &0 &0 &0 \\0 &0 &0 &0 &0   \\0 &0  &-3 &4 &-1 \\ 0 &0 &0 &0 &0  \\ 0 &0 &0 &0 &0 \end{pmatrix}, \qquad q_{02}=\begin{pmatrix} 0 &0 &0 &0 &0 \\0 &0 &0 &0 &0   \\0 &1  &-2 &1 & 0 \\ 0 &0 &0 &0 &0  \\ 0 &0 &0 &0 &0 \end{pmatrix}.
$$
We compute $\frac{\partial u}{\partial x}:=u_{01}=q_{01} \otimes u$, $\frac{\partial^2 u}{\partial x^2}:=u_{02}=q_{02} \otimes u$. The sub-indices are used to specify the target derivative, e.g., $u_{ij}$ represents $\frac{\partial^{i+j}u}{\partial x^j\partial y^i}$ and it can be computed via $q_{ij} \otimes u$. This terminology can be easily generalized to higher-dimensional functions and their derivatives. Therefore, learning a derivative to be applied to a function in the governing equation is reduced to learning a matrix representing the corresponding convolution kernel. The degree of freedom of a convolution kernel matrix is large, e.g., for the $5$ by $5$ matrix above, the degree of freedom is $25$. Optimizing $25$ continuous variables to obtain the best convolution kernel for one derivative is also expensive. Therefore, in FEX for governing equations, a set of pre-determined filters (e.g., $\{q_{00},q_{01},q_{02},q_{10},q_{20},q_{11}\}$) is determined with a user-specified discretization scheme. To choose an appropriate derivative in the governing equation, we optimize $a_{00}q_{00}+a_{01}q_{01}+a_{02}q_{02}+a_{10}q_{10}+a_{20}q_{20}+a_{11}q_{11}$ over $6$ continuous variables $a_{00},a_{01},a_{02},a_{10},a_{20}$ and $a_{11}$. The derivative will be determined according to the dominant variable. See the discussion in Appendix~\ref{trainable convolution filters} for more details. 

\begin{figure*}[htbp]
    \centerline{\includegraphics[width=5in]{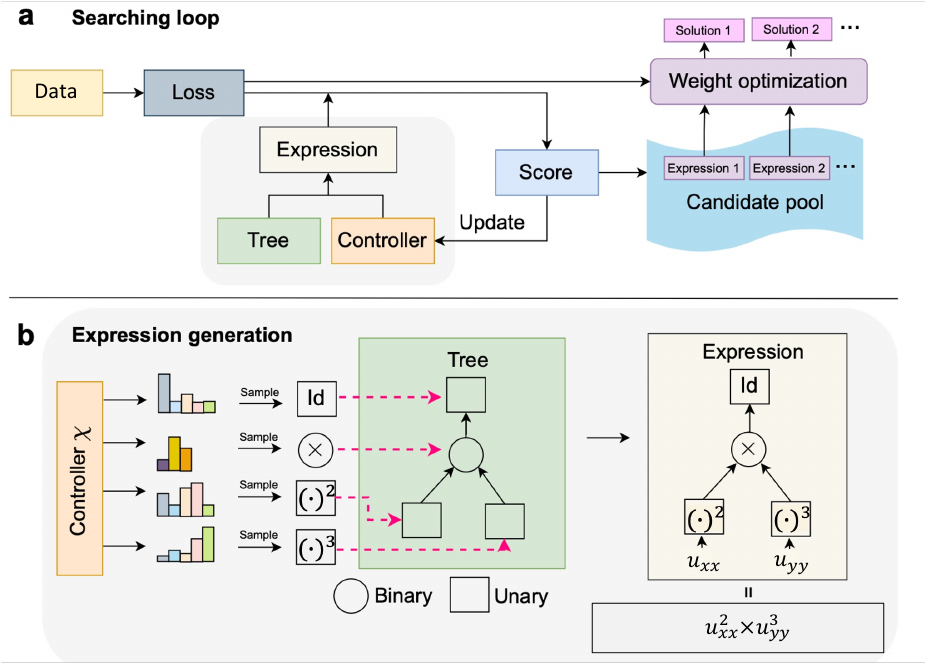}}
    \begin{center}
    \caption{The representation of the major components of our FEX. (a) The searching loop for the symbolic solution consists of expression generation, score computation, controller update, and candidate optimization. (b) The depiction of the expression generation with a binary tree and a controller $\chi$. The appropriate derivatives of $u$ at the leaf level of the binary tree will be learned via continuous optimization.
    } 
    \label{schematic digram}
    \end{center}
\end{figure*}

\subsection{Finite Expression Methods for Discovering Physical Laws}

The finite expression method   \cite{liang2022finite} is a deep reinforcement learning technique to approximate solutions of PDEs within the function space composed of a finite number of analytic expressions.   FEX \cite{liang2022finite} can achieve high accuracy, including machine accuracy, with a memory complexity polynomial in the problem dimension and a manageable time complexity for solving high-dimensional PDEs.  

In this paper, we employ the idea of FEX to learn symbolic governing equations from data. The architecture of our FEX is illustrated in Figure ~\ref{schematic digram}. It is important to note that our FEX essentially differs from that in \cite{liang2022finite}. While Figure 3 in \cite{liang2022finite} selects differential operators during discrete optimization of the operator list in the computational tree, we compute derivatives in advance and select them through continuous optimization as introduced in Section \ref{derivatives computing}. 
In our FEX method, we first introduce mathematical expressions in Section~\ref{function space}. Next, our combinatorial optimization  (CO) is formulated for the parameter and operator selection to seek the expression that approximates the governing equation. RL is employed to select operators. Particularly, a controller parametrized as a deep neural network acts as the RL agent that takes actions according to a policy as a probability distribution function as in the original FEX method in \cite{liang2022finite}. The action contains a list of operators to be filled in a binary tree (See in Figure ~\ref{schematic digram}b). The binary tree with these operators still has trainable parameters to formulate a final mathematical expression. The state of the RL is the best mathematical expression with the best set of trainable parameters (specified via continuous optimization) such that the expression is the best governing equation corresponding to the chosen operators in the binary tree.  A functional $\mathcal{L}$ will be designed to quantify how well an expression governs the dynamic system, which provides a loss function to optimize trainable parameters of a binary tree and the reward in RL for optimizing operator selection. A policy gradient method will be applied to solve the RL optimization.

\subsubsection{Expression spaces in FEX}\label{function space}
We shall first review the definition of a mathematical expression as proposed in \cite{liang2022finite}. Next, we shall introduce FEX for discovering physical laws.

\textsc{Definition 1} (Mathematical expressions). \cite{liang2022finite} \textit{A mathematical expression is a valid function formulated through a combination of symbols, following well-formed syntax and rules. The symbols encompass operands, which include variables and numbers, as well as operators such as addition ('+'), subtraction ('-'), integration, differentiation, brackets, and punctuation.}

In this paper, given a dictionary of operators, $\mathbb{S}_k$ denotes the space of mathematical expressions of equations with at most $k$ operators in the dictionary.

\textsc{Definition 2} (FEX for discovering physical laws). \textit{FEX is a methodology to learn a symbolic governing equation by seeking an expression with finitely many operators to approximate the target equation. This approach involves the search for a concise mathematical expression, typically of limited complexity and length, which can effectively capture and represent the behavior or relationship described by the target equation.}

\subsubsection{Identify the expressions of governing equations in FEX}\label{solution expression}

In the context of learning a finite-expression equation (i.e., the number of operators in this equation is a finite number, say, at most $k$), the problem can be formulated as a combinatorial optimization problem over the expression space $\mathbb{S}_k$. The objective is to find an equation within $\mathbb{S}_k$ that minimizes a problem-dependent functional $\mathcal{L}$, which quantifies how well an expression governs the unknown dynamic system or PDE. Mathematically, the problem can be stated as follows:
$$
\min_{g \in \mathbb{S}_k} \mathcal{L}(g),
$$
where $g$ represents the finite-expression equation within $\mathbb{S}_k$.

After the reformulation into a combinatorial optimization problem, researchers can leverage various optimization algorithms and techniques to search for the finite-expression equation that best approximates the unknown PDE or dynamic system. The restriction to the expression space $\mathbb{S}_k$, which consists of equations with a finite number of operators, allows for obtaining a compact and interpretable representation of the governing equation while capturing the essential behavior or relationship of the system.

\subsubsection{Binary trees in FEX}\label{binarytree}

\begin{figure*}[htbp]
    \centerline{\includegraphics[width=6.5in]{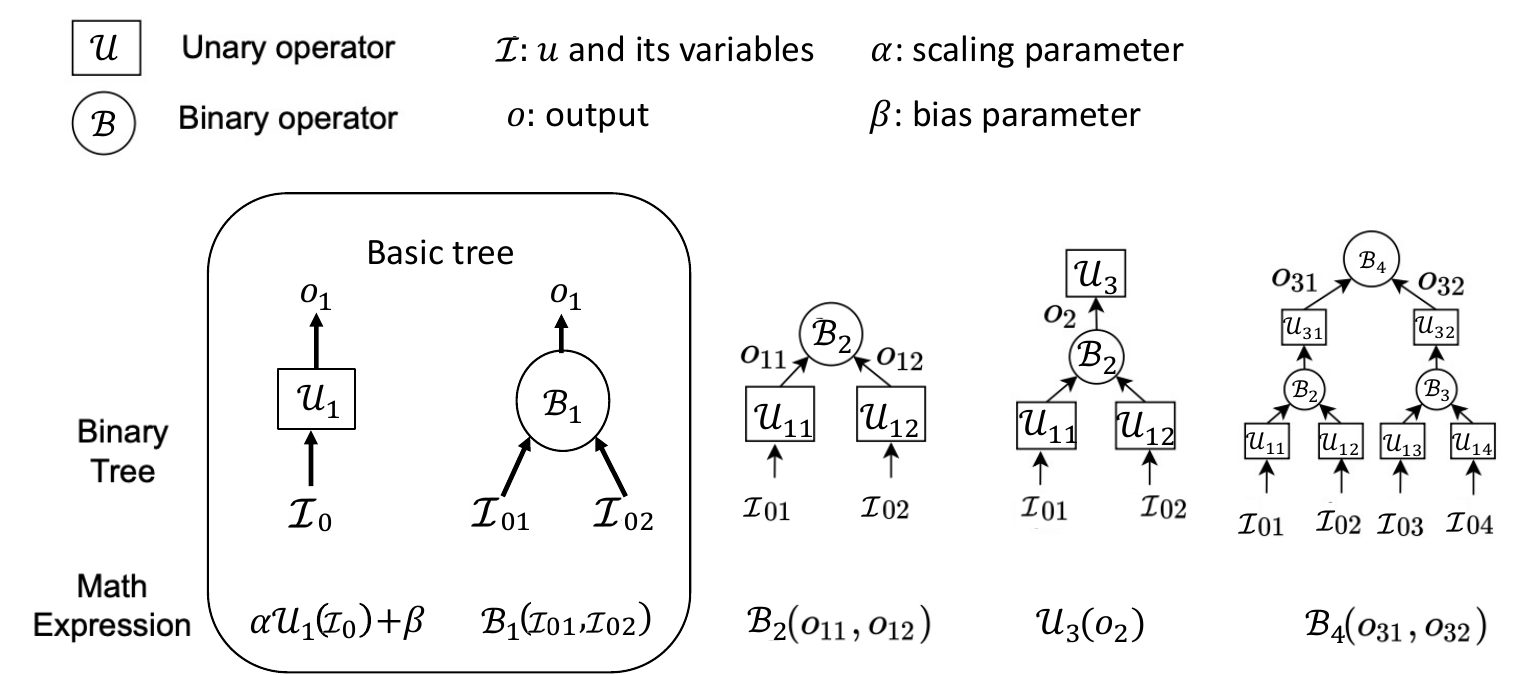}}
    \begin{center}
    \caption{The binary tree structure in the fully combinatorial optimization scheme. Only first-order differential operators are included in the dictionary of unary operators. Given a prescribed dictionary of unary and binary operators, e.g., $\{+,-,\times,/, \sin, \cos, \exp, \partial_x,\partial_y,\dots\}$, when the inputs are chosen as a function $u$ and its variables, complex governing equations of $u$ with possibly high-order derivatives and varying coefficients can be generated via operator compositions through a deep binary tree.}
    \label{binary tree_full CO}
    \end{center}
\end{figure*}

\begin{figure*}[htbp]
    \centerline{\includegraphics[width=6.5in]{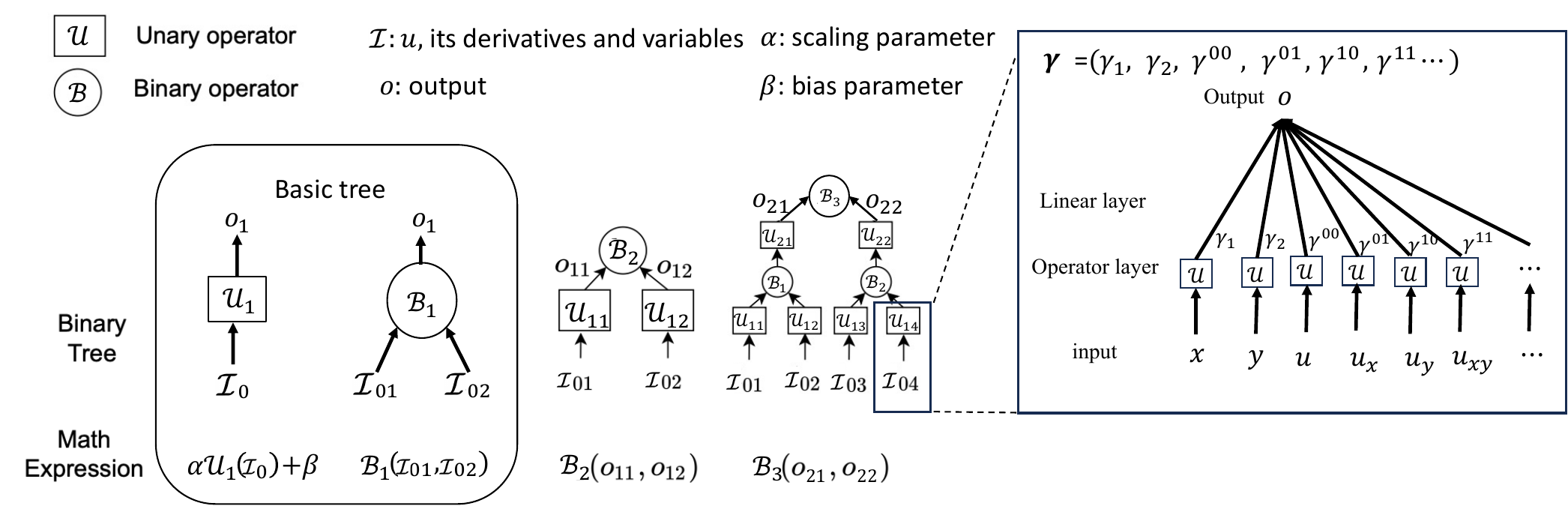}}
    \begin{center}
    \caption{The binary tree structure in the partially relaxed combinatorial optimization for learning derivatives. Instead of including differential operators in the operator dictionary, leaf nodes are designed to learn to select appropriate derivatives of $u$ to form a governing equation via continuous optimization.}
    \label{binary tree}
    \end{center}
\end{figure*}

In the FEX framework, a finite expression is formulated using a binary tree $\mathcal{T}$ commonly adopted in computer algebra. Each node of the binary tree is associated with a unary or a binary operator (possibly with a few trainable parameters) selected from a prescribed dictionary of operators. In FEX, the inorder traversal of binary trees is adopted to list the operators of the tree into a vector of operators denoted as $\boldsymbol{e}$, which is also called an operator sequence. The inputs of the leaf nodes consist of the function $u$ (or its discretization) satisfying the unknown governing equation and possibly together with the derivatives of $u$.  Each tree node not at the leaf level takes the outputs of its children nodes as inputs and generates an output by evaluating its associated operator. The generated output will serve as the input of its parent node. The output of the root node represents a mathematical expression generated by the whole tree as a governing equation of $u$. To enhance the expressiveness of binary trees, we introduce $\alpha$ and $\beta$ as trainable scaling and bias parameters, respectively, in unary operators. For example, the output $o$ of a unary operator $\mathcal{U}$ with an input $\mathcal{I}$ is computed via $o=\alpha \mathcal{U}(\mathcal{I})+\beta$.  Let $\boldsymbol{\theta}$ denote the set of all trainable parameters of the binary tree $\mathcal{T}$, then the mathematical expression generated with selected operators in $\boldsymbol{e}$ and parameters in $\boldsymbol{\theta}$ is denoted as $g(u ; \mathcal{T}, \boldsymbol{e}, \boldsymbol{\theta})$ as a functional or operator of $u$. In FEX, there are two types of binary trees to generate a governing equation, and their examples are given in Figures~\ref{binary tree_full CO} and \ref{binary tree} corresponding to two different optimization methods to learn derivatives, respectively.  

The first type of binary trees follows the idea in the original FEX paper \cite{liang2022finite} when the unary operator set contains basic first-order derivatives (e.g., $\frac{\partial}{\partial x}$ and $\frac{\partial}{\partial y}$ for a two-dimensional problem) without operators for higher order derivatives. As illustrated in Figure~\ref{binary tree_full CO}, the main idea is to generate a complex governing equation via the compositional computation through a binary tree with the most basic operators. Each tree level contains only unary operators or binary operators. Unary operators and binary operators are applied alternatively at different tree levels. Unary operators at the leaf level take an input $\mathcal{I}$ as the function $u$ or one of its variables and the output of the binary operator at the roof level generates a governing equation $g$ of $u$ with possible varying coefficients. For example, the mathematical expression $\frac{\partial u}{\partial x}+\frac{\partial u}{\partial y}$ can be generated by a binary tree with two levels, e.g., the tree $\mathcal{B}_2(o_{11},o_{12})$ in Figure~\ref{binary tree_full CO} with $\mathcal{B}_2$ as ``$+$", where $o_{11}= \alpha_1 \mathcal{U}_{11}(\mathcal{I}_{01}) + \beta_1$ with $\mathcal{U}_{11} = \frac{\partial}{\partial x}$, $\mathcal{I}_{01}=u$, $\alpha_1=1$, and $\beta_1=0$; $o_{12}= \alpha_2 \mathcal{U}_{12}(\mathcal{I}_{02})+\beta_2$ with $\mathcal{U}_{12} = \frac{\partial}{\partial y}$, $\mathcal{I}_{02}=u$, $\alpha_2=1$, and $\beta_2=0$. For another example, a mathematical expression with higher order derivatives and varying coefficients $\sin(x)\frac{\partial^2 u}{\partial x^2} + \cos(x)\frac{\partial u}{\partial x}$ can be generated by a binary tree with three levels, e.g. $\mathcal{U}_3(o_2)$ in Figure~\ref{binary tree_full CO} with $\mathcal{U}_3=\frac{\partial}{\partial x}$ and $o_2 =\mathcal{B}_2(o_{11},o_{12}):=o_{11}\times o_{12}$, where $o_{11}= \mathcal{U}_{11}(\mathcal{I}_{01}):= \frac{\partial u}{\partial x}$ similarly to the previous example, and $o_{12}= \alpha_2 \mathcal{U}_{12}(\mathcal{I}_{02})+\beta_2:= \sin(x)$ with $\mathcal{U}_{12}=\sin$, $\alpha_2=1$, $\beta_2=0$, and $\mathcal{I}_{02}=x$. Therefore, even if the operator dictionary only contains first-order derivatives, high-order derivatives can be learned via combinatorial optimization to select appropriate first-order derivatives in tree nodes, which can generate complex expressions via operator compositions in the tree structure without prescribed high-order derivatives in the operator dictionary. Therefore, the learning algorithm associated with this type of binary trees is called a fully combinatorial optimization (FCO) scheme. The optimization goal is to find the most appropriate operator sequence $\boldsymbol{e}$ and trainable parameters $\boldsymbol{\theta}$ consisting of all scaling and bias parameters such that the expression $g(u ; \mathcal{T}, \boldsymbol{e}, \boldsymbol{\theta})$ best governs the given function $u$. Though the FCO provides flexibility to generate all possible complex governing equations with high-order derivatives, the corresponding combinatorial optimization for selecting first-order derivatives is challenging. Relaxing the combinatorial optimization for learning derivatives into continuous optimization would lead to a more efficient way to learn governing equations, which motivates a new design of binary trees in this paper as follows.

In the second type of binary trees, as illustrated in Figure~\ref{binary tree}, the main tree structure is the same as the first type in Figure \ref{binary tree_full CO}, but there is no differential operator in the operator dictionary. Furthermore, the input of a leaf node with a unary operator is denoted as $\mathcal{I}$ consisting of $u$, prescribed derivatives of $u$, and the variables of $u$. The unary operator at a leaf node, $\mathcal{U}$, is applied entrywise to the input vector $\mathcal{I}$, the results of which are added together to form an output $o$ with learnable linear combination coefficients stored in a vector $\boldsymbol{\gamma}$ as illustrated in Figure~\ref{binary tree} (right). Recall that the derivatives of $u$ can be computed with their corresponding convolution filters. For example, as introduced in Section \ref{derivatives computing}, in the case of a two-dimensional function $u(x,y)$, $u_{ij}$ represents $\frac{\partial^{i+j}u}{\partial x^j\partial y^i}$ and it can be computed via $q_{ij} \otimes u$, where $q_{ij}$ is a prescribed convolution filter. The input to a leaf node $\mathcal{I}$ is then equal to $\{x,y,q_{00} \otimes u,q_{01} \otimes u, q_{10} \otimes u,\dots\}$. The output of a leaf node is then defined as
\[
o = \gamma_{1} \mathcal{U}(x)+ \gamma_{2}\mathcal{U}(y)+\sum \gamma^{ij}\mathcal{U}(q_{ij} \otimes u),
\]
and $\boldsymbol{\gamma}$ is used to denote the coefficient vector $(\gamma_{1},\gamma_{2},\gamma^{00},\gamma^{01},\gamma^{10},\gamma^{11},\dots)$. Unary operators not at the leaf level are applied in the same manner as in the first type of binary trees with scaling and bias parameters $\alpha$ and $\beta$, respectively. Let $\boldsymbol{\theta}$ be the set of all trainable parameters including all linear combination coefficients $\boldsymbol{\gamma}$, scaling parameters $\alpha$, and bias parameters $\beta$ in all nodes. The optimization goal is to find the most appropriate operator sequence $\boldsymbol{e}$ and trainable parameters $\boldsymbol{\theta}$ such that the expression $g(u ; \mathcal{T}, \boldsymbol{e}, \boldsymbol{\theta})$ best governs the given function $u$. This optimization method has relaxed the combinatorial optimization for selecting derivatives to continuous optimization over $\boldsymbol{\gamma}$. Therefore, this method is referred to as the partially relaxed combinatorial optimization (PCO) for learning derivatives. This PCO offers computational advantages and ease of
implementation compared to the FCO. In this paper, the PCO scheme is adopted in our numerical experiments for simplicity and efficiency.

In practice, the computational cost of FEX is exponentially large in the depth of binary trees. Applying prior knowledge of the physical systems to reduce the depth of binary trees can significantly improve the computational efficiency of FEX. For example, if the unknown governing equation of $u$ is a polynomial of $u$ and its derivatives, it is only necessary to apply unary operators at the leaf level and no unary operator is applied above the leaf level. This can essentially reduce the tree depth by half. In most physical systems, high-order derivatives rarely appear and, hence, a short input vector $\mathcal{I}$ with low-order derivatives of $u$ is sufficient in most applications, which can also save the computational cost efficiently.

\subsubsection{Implementation of FEX}\label{Implementation of FEX}

Given a fixed binary tree $\mathcal{T}$ with $k_{\mathcal{T}}$ nodes, the FEX framework aims to identify the governing equation from the expression space generated by $\mathcal{T}$, which is a subset of $\mathbb{S}_{k_{\mathcal{T}}}$. With the abuse of notations, we still use $\mathbb{S}_{k_{\mathcal{T}}}$ to denote the expression space generated by $\mathcal{T}$. Given $u$ (or equivalently its observations) and a fixed binary tree $\mathcal{T}$, the optimal mathematical expression for the governing equation of $u$ in $\mathbb{S}_{k_{\mathcal{T}}}$ is obtained by solving the following minimization problem:
$$
\min_{\boldsymbol{e}, \boldsymbol{\theta}} \mathcal{L}(g(u ; \mathcal{T}, \boldsymbol{e}, \boldsymbol{\theta})),
$$
where $\boldsymbol{e}$ represents the operator sequence, $\boldsymbol{\theta}$ denotes the set of all learnable parameters in the binary tree $\mathcal{T}$, $g(u; \mathcal{T}, \boldsymbol{e}, \boldsymbol{\theta})$ represents the finite-expression equation of $u$ generated by the binary tree $\mathcal{T}$. Now we present the FEX framework in detail to solve the above minimization problem. The main idea is to minimize the loss function $\mathcal{L}$ alternatively over $\boldsymbol{e}$ via RL (e.g., the risk-seeking policy gradient method proposed in~\cite{tamar2014policy}) and over $\boldsymbol{\theta}$ via gradient-based methods (e.g., Adam~\cite{kingma2014adam} and the Broyden-Fletcher-Goldfarb-Shanno method (BFGS)~\cite{fletcher2013practical}). Once a good set of operators in $\boldsymbol{e}$ has been chosen, optimizing $\boldsymbol{\theta}$ is relatively simple and efficient. The main optimization algorithm is illustrated in Figure~\ref{schematic digram} and presented in Algorithm~\ref{algo}. Our framework consists of a searching loop, as depicted in Figure~\ref{schematic digram}a, which encompasses four key components:
 
\begin{itemize}
\item Reward computation: In this step, rewards for choosing actions in RL are computed. A higher reward assigned to an operator sequence $\boldsymbol{e}$ indicates a greater likelihood of discovering the governing equation accurately.
\item Operator sequence generation: A controller is employed in the RL process to generate operator sequences. This step involves taking actions in RL to explore various operator sequences.
\item Controller update: The controller is updated to increase the probability of generating improved operator sequences with higher rewards. To achieve this, the policy gradient method is utilized to update the controller and optimize the policy function within the RL framework.
\item Candidate optimization: During the controller update process, FEX maintains a candidate pool to store operator sequences that exhibit relatively high rewards. Subsequently, the parameters $\boldsymbol{\theta}$ of these high-reward operator sequences are optimized to approximate the governing equation more effectively. This approach employs a policy-gradient-based updating method to explore a wider range of potential solutions.
\end{itemize}
These four components will be introduced in detail in the following subsections. By integrating these four components, the FEX framework as summarized in Algorithm~\ref{algo} provides a systematic approach to identify the optimal operator sequence and trainable parameters to generate a governing equation that best fits given observations of $u$. In Algorithm~\ref{algo}, the tree depth is fixed and denoted as $D$. In practice, a smaller tree depth is more efficient in computation but may not be sufficient to generate a good governing equation to fit data. Therefore, it is better to gradually expand the tree size from a smaller one to a larger one if a smaller one is not enough. This idea can be summarized in Algorithm~\ref{algo: expanding trees} with the process of expanding a tree set $\left\{\mathcal{T}_1, \mathcal{T}_2, \cdots\right\}$ for searching the governing equation to reach a target accuracy.

\begin{algorithm}[htbp]
   \caption{FEX for Discovering Physical Laws}
   \label{algo}
\begin{algorithmic}
   \STATE {\bfseries Input:} The functional $\mathcal{L}$ associated with derivatives of the solutions of PDEs or dynamics systems; A tree $\mathcal{T}$ of a fixed depth $D$; Searching loop iteration $T$; Coarse-tune iteration $T_1$ with Adam; Coarse-tune iteration $T_2$ with BFGS; Fine-tune iteration $T_3$ with Adam; Pool size $K$; Batch size $N$.\\
   {\bfseries Output:} The governing equation $g(u ; \mathcal{T}, \hat{\boldsymbol{e}}, \hat{\boldsymbol{\theta}})$.\\
   \STATE Initialize agent $\chi$ for the tree $\mathcal{T}$
   \STATE $\mathbb{P} \leftarrow\{\}$
   \FOR{$\_$ from 1 {\bfseries to} $T$}
   \STATE Sample $N$ sequences $\left\{\boldsymbol{e}^{(1)}, \boldsymbol{e}^{(2)}, \cdots, \boldsymbol{e}^{(N)}\right\}$ from $\boldsymbol{\chi}$
   \FOR{$n$ from 1 {\bfseries to} $N$}
   \STATE Optimize $\mathcal{L}\left(g\left(u ; \mathcal{T}, \boldsymbol{e}^{(n)}, \boldsymbol{\theta}\right)\right)$ by coarse-tune with $T_1+T_2$ iterations.
   \STATE Compute the reward $R\left(\boldsymbol{e}^{(n)}\right)$ of $\boldsymbol{e}^{(n)}$
   \IF{$\boldsymbol{e}^{(n)}$ belongs to the top- $K$ of $S$}
   \STATE $\mathbb{P}$.append $\left(\boldsymbol{e}^{(n)}\right)$
   \STATE $\mathbb{P}$ pops some $\boldsymbol{e}$ with the smallest reward when overloading
   \ENDIF
   \ENDFOR
   \STATE Update $\boldsymbol{\chi}$ according to Section~\ref{objective function}
   \ENDFOR
   \FOR{$\boldsymbol{e}$ in $\mathbb{P}$}
   \STATE Fine-tune $\mathcal{L}(g(u ; \mathcal{T}, \boldsymbol{e}, \boldsymbol{\theta}))$ with $T_3$ iterations.
   \ENDFOR
\end{algorithmic}
\end{algorithm}

\begin{algorithm}[htbp]
   \caption{FEX with progressively expanding trees}
   \label{algo: expanding trees}
\begin{algorithmic}
    \STATE {\bfseries Input:}Tree set $\left\{\mathcal{T}_1, \mathcal{T}_2, \cdots\right\} ;$ Error tolerance $\epsilon$;\\
    {\bfseries Output:}the solution $g(u ; \hat{\mathcal{T}}, \tilde{\boldsymbol{e}}, \tilde{\boldsymbol{\theta}})$.
    \FOR{$\mathcal{T}$ in $\left\{\mathcal{T}_1, \mathcal{T}_2, \cdots\right\}$}
    \STATE Initialize the agent $\boldsymbol{\chi}$ for the tree $\mathcal{T}$
    \STATE Obtain $g(u ; \mathcal{T}, \hat{\boldsymbol{e}}, \boldsymbol{\theta})$ from Algorithm~\ref{algo}
    \IF{$\mathcal{L}(g(\cdot ; \mathcal{T}, \hat{\boldsymbol{e}}, \hat{\boldsymbol{\theta}})) \leq \epsilon$}
    \STATE Break
    \ENDIF
    \ENDFOR
    \STATE {\bfseries return} {the expression with smallest functional value.}
\end{algorithmic}
\end{algorithm}

\subsubsection{Reward computation}\label{reward compute}

The reward assigned to an operator sequence $\boldsymbol{e}$, denoted as $R(\boldsymbol{e})$, plays a vital role in guiding the optimization of the controller within the FEX framework. It aims to generate improved operator sequences and maintain a candidate pool consisting of high-scoring solutions. The reward in FEX is defined as
$$
R(\boldsymbol{e}):=(1+L(\boldsymbol{e}))^{-1},
$$
where $L(\boldsymbol{e})=\min_{\boldsymbol{\theta}} \mathcal{L}(g(u ; \mathcal{T}, \boldsymbol{e}, \boldsymbol{\theta})) $. The reward $R(\boldsymbol{e})$ is bounded between 0 and 1. As $L(\boldsymbol{e})$ approaches 0, indicating a closer approximation to the true governing equation, the reward $R(\boldsymbol{e})$ tends to 1, reflecting a higher-quality operator sequence. Conversely, larger values of $L(\boldsymbol{e})$ indicate a greater discrepancy from the true governing equation, resulting in a lower reward.

 To expedite the computation of the reward, a combination of first-order and second-order optimization algorithms is employed sequentially. Initially, a first-order algorithm such as stochastic gradient descent~\cite{rumelhart1986learning} or Adam~\cite{kingma2014adam} is employed. However, optimizing $L(\boldsymbol{e})$ using a first-order algorithm can be time-consuming due to the need for a large number of iterations. 
To overcome this issue, a second-order optimization algorithm such as the BFGS~\cite{fletcher2013practical} is incorporated. The second-order algorithm takes advantage of the Hessian matrix to achieve faster convergence when provided with a good initial guess. In the proposed approach, the first-order algorithm is first employed for $T_1$ steps to obtain a robust initial guess. Subsequently, the second-order algorithm, benefiting from the obtained good initial guess, is used for an additional $T_2$ steps. By sequentially utilizing the strengths of both algorithms, the efficiency of optimizing $L(\boldsymbol{e})$ is enhanced, leading to accelerated computation of the reward $R(\boldsymbol{e})$.

\subsubsection{Operator sequence generation}\label{operation generation}
The proposed controller, denoted as $\boldsymbol{\chi}_{\Phi}$ with parameters in $\Phi$, serves as the policy function within the RL framework to generate operator sequences $\boldsymbol{e}$ that yield high rewards. By updating the parameter $\Phi$, we optimize $\boldsymbol{\chi}_{\Phi}$ to increase the probability of generating good operator sequences.
In our approach, the controller $\boldsymbol{\chi}_{\Phi}$ treats the tree node values of $\mathcal{T}$ as random variables. It outputs probability mass functions $ \boldsymbol{p}^1_{\Phi}, \boldsymbol{p}^2_{\Phi}, \cdots, \boldsymbol{p}^k_{\Phi} $ that characterize the distributions of these random variables, where $k$ represents the total number of nodes in the tree.
Based on the probability mass functions obtained from $\boldsymbol{\chi}_{\Phi}$, an operator sequence $\boldsymbol{e}$ is generated by randomly sampling the distributions. Specifically, $\boldsymbol{e}$ is composed of individual elements $\boldsymbol{e}_1, \boldsymbol{e}_2, \cdots, \boldsymbol{e}_k$, where each $\boldsymbol{e}_i$ corresponds to an operator sampled from the probability mass function associated with the respective tree node. This sampling process enables the generation of diverse operator sequences from $\boldsymbol{\chi}_{\Phi}$, as illustrated in Figure~\ref{schematic digram}.

\subsubsection{Controller update}\label{controller update}

In our approach, the controller $\boldsymbol{\chi}_{\Phi}$ is implemented as a neural network, and its parameters are denoted as $\Phi$. The objective of training the controller is to maximize the expected reward associated with a sampled operator sequence $\boldsymbol{e}$. Mathematically, this objective is expressed as:\begin{equation*}
\mathcal{J}(\Phi) := \mathbb{E}_{\boldsymbol{e} \sim \boldsymbol{\chi}_{\Phi}} [R(\boldsymbol{e})],
\label{objective function}
\end{equation*}
where $\boldsymbol{e} \sim \boldsymbol{\chi}_{\Phi}$ indicates the process of sampling an operator sequence $\boldsymbol{e}$ from the controller $\boldsymbol{\chi}_{\Phi}$, and $R(\boldsymbol{e})$ represents the reward associated with the sampled operator sequence $\boldsymbol{e}$.

To optimize the controller $\boldsymbol{\chi}_{\Phi}$, we employ a policy-gradient-based updating method in RL. The policy gradient $\nabla_{\Phi} \mathcal{J}(\Phi)$ is computed as follows:
$$
\nabla_{\Phi} \mathcal{J}(\Phi)=\mathbb{E}_{\boldsymbol{e} \sim \chi_{\Phi}}\left\{R(\boldsymbol{e}) \sum_{i=1}^k \nabla_{\Phi} \log \left(\boldsymbol{p}_{\Phi}^i\left(e_i\right)\right)\right\},
$$where $\boldsymbol{p}_{\Phi}^i\left(e_i\right)$ represents the probability of selecting the operator $e_i$ at the $i$-th position of the sequence. In practice, we compute an approximation of this gradient using a batch of $N$ sampled operator sequences ${\boldsymbol{e}^{(1)}, \boldsymbol{e}^{(2)}, \ldots, \boldsymbol{e}^{(N)}}$ as follows:
$$
\nabla_{\Phi} \mathcal{J}(\Phi) \approx \frac{1}{N} \sum_{j=1}^N\left\{R\left(\boldsymbol{e}^{(j)}\right) \sum_{i=1}^k \nabla_{\Phi} \log \left(\boldsymbol{p}_{\Phi}^i\left(e_i^{(j)}\right)\right)\right\}.
$$
To update the parameter $\Phi$ of the controller, we use the gradient ascent method with a learning rate $\eta$:
$$
\Phi \leftarrow \Phi+\eta \nabla_{\Phi} \mathcal{J}(\Phi).
$$The objective function $\mathcal{J}(\Phi)$ aims to improve the average reward of the $N$ sampled operator sequences. However, to increase the probability of obtaining the best equation expression, we incorporate the objective function proposed in~\cite{petersen2019deep} and the risk-seeking policy gradient method proposed in~\cite{tamar2014policy}. The modified objective function becomes:
$$
\mathcal{J}(\Phi)=\mathbb{E}_{\boldsymbol{e} \sim \chi_{\Phi}}\left\{R(\boldsymbol{e}) \mid R(\boldsymbol{e}) \geq R_{\nu, \Phi}\right\},
$$
where $R_{\nu, \Phi}$ represents the $(1-\nu) \times 100 \%$-quantile of the reward distribution generated by $\chi_{\Phi}$ and $\nu \in [0,1]$.   Therefore, the gradient computation is updated as:
\begin{align*}
\nabla_{\Phi} \mathcal{J}(\Phi) \approx &\frac{1}{N} \sum_{j=1}^N\{\left(R\left(\boldsymbol{e}^{(j)}\right)-\hat{R}_{\nu, \Phi}\right) \mathbf{1}_{\left\{R\left(\boldsymbol{e}^{(j)}\right) \geq \hat{R}_{\nu, \Phi}\right\}} \sum_{i=1}^k \nabla_{\Phi} \log \left(\boldsymbol{p}_{\Phi}^i\left(e_i^{(j)}\right)\right) \},
\end{align*}
where the indicator function $\mathbf{1}$ is used as a binary indicator that returns 1 when the condition $R\left(\boldsymbol{e}^{(j)}\right) \geq \hat{R}_{\nu, \Phi}$ is satisfied and 0 otherwise. The value $\hat{R}_{\nu, \Phi}$ represents the $(1-\nu)$-quantile of the reward distribution generated by the controller $\boldsymbol{\chi}_\Phi$. It is computed based on the rewards $R\left(\boldsymbol{e}^{(i)}\right)$ for each sampled operator sequence $\boldsymbol{e}^{(i)}$ where $i = 1, \ldots, N$.

\subsubsection{Candidate optimization}\label{candidate optimization}
For the sake of not missing good operator sequences, a candidate pool $\mathbb{P}$ with capacity $K$ is maintained to store several high scores $\boldsymbol{e}$. During the search loop, if the size of $\mathbb{P}$ is less than the capacity $K, \boldsymbol{e}$ will be put in $\mathbb{P}$. If the size of $\mathbb{P}$ reaches $K$ and $R(\boldsymbol{e})$ is larger than the smallest reward in $\mathbb{P}$, then $\boldsymbol{e}$ will be appended to $\mathbb{P}$ and the one with the least reward will be removed. After the searching loop, for every $\boldsymbol{e} \in \mathbb{P}$, the objective function $\mathcal{L}(g(u ; \mathcal{T}, \boldsymbol{e}, \boldsymbol{\theta}))$ is optimized over $\boldsymbol{\theta}$ using a first-order algorithm like Adam with a small learning rate for $T_3$ iterations.

\section{Numerical Results}
In this section, we present the numerical results that demonstrate the accuracy and effectiveness of our FEX algorithm in various scenarios. We evaluate the performance of our FEX in discovering PDEs, nonlinear dynamical systems, and nonlinear functions. We compare our FEX with other existing algorithms, such as PDE-Net 2.0~\cite{long2019pde}, SINDy~\cite{brunton2016discovering}, generic programming (GP)~\cite{koza1994genetic} and the symbolic physics learner (SPL)~\cite{sun2023symbolic}.
All experiments are conducted on a Macbook Pro equipped with an Apple M1 Pro chip and 16 GB RAM. 

  
  
  


 \textbf{Experimental Settings:} 
In the implementation of FEX, several key numerical choices are made in each of the three main parts:

 (1) \textit{Reward computation.} The reward computation involves updating the functional using two different optimization algorithms, Adam and BFGS. The update is performed with the iteration number $T_1 = 500$ for Adam and $T_2 = 20$ for BFGS, respectively. These choices determine the convergence criteria for the optimization process and can be adjusted based on the specific problem and desired accuracy. The combination of Adam and BFGS optimizers allows for efficient convergence and improved performance in finding the optimal mathematical expression.
 
(2) \textit{Operator sequence generation.} To generate mathematical expressions, a binary tree of depth $4$ consisting of six unary operators and five binary operators is employed. For simplicity and efficiency, the unary operators are added only at the leaf level of the binary tree. The binary set $\mathbb{B} = \{+,-,\times\}$ represents the available binary operators, while the unary set $\mathbb{U} = \{1, \mathrm{Id}, (\cdot)^2, \ldots\}$ represents the available unary operators. These sets define the range of operators that can be used to form mathematical expressions. Additionally, a fully connected neural network (NN) is utilized as a controller $\chi_{\Phi}$, with a constant input arbitrarily chosen by user. The output size of the controller NN is $n_1 + n_2$, where $n_1 = 5$ and $n_2 = 6$ represent the numbers of binary and unary operators, respectively. This design allows the controller to generate operator sequences based on the given inputs and the defined sets of operators.

(3) \textit{Controller update.} The controller update involves training the controller using a policy gradient update. In this process, a batch size of $N = 5000$ is used. The controller is trained for a total of 500 iterations.  The policy gradient update enables the controller to improve its ability to generate operator sequences which lead to higher rewards, enhancing the overall performance of our FEX algorithm.

 
The objective of FEX is to discover the governing PDEs from a series of measured physical quantity $\{u(t, x, y)\}$, where $u$ represents the state of the system, and $t$, $x$, and $y$ are time and spatial coordinates, respectively. It is easy to generate our discussion to functions with more variables. We assume that the observed data are associated with a PDE in the general form:
$$
    u_t(t, x, y)=F\left(u, u_x, u_y, u_{x x}, u_{x y}, u_{y y}, \ldots\right),
$$
where $u(t, \cdot): \Omega \mapsto \mathbb{R}^d, F\left(u, u_x, u_y, u_{x x}, u_{x y}, u_{y y}, \ldots\right) \in \mathbb{R}^d,(x, y) \in \Omega \subset \mathbb{R}^2, t \in[0, T]$. To achieve this objective, we utilize the FEX algorithm to design a feed-forward NN capable of discovering the unknown PDE from solution samples. The FEX algorithm can search for concise mathematical expressions within a function space of limited complexity to unveil the analytic form of the response function $F$ and identify the specific differential operators involved in the underlying PDE. Once the NN has been trained and the PDE has been discovered, we can leverage this information for long-term predictions of the system's dynamic behavior. Provided with any given initial condition, the trained NN can simulate the system's evolution over time, enabling us to make predictions about its future state. This capability is valuable for understanding the long-term behavior of the equation and assessing its stability, sensitivity to initial conditions, or response to external inputs.

In our numerical tests, we employ the forward Euler method as the temporal discretization scheme for evolving the PDE in time. The forward Euler method is used to approximate the solution at the next time step, denoted as $\tilde{u}(t+\delta t, \cdot)$, based on the current solution $\tilde{u}(t, \cdot)$. More precisely, we use the following expression to approximate the solution at the next time step:
$$
\tilde{u}(t+\delta t, \cdot) \approx \tilde{u}(t, \cdot)+\delta t \cdot \operatorname{FEXTree}\left(D_{00} \tilde{u}, D_{01} \tilde{u}, \cdots\right),
$$
where $\tilde{u}(t, \cdot)$ represents the current approximation of the solution at time $t$,   $\tilde{u}(t+\delta t, \cdot)$ is the predicted value at the next time step $t+\delta t$,   $\operatorname{FEXTree}$ refers to the binary tree $\mathcal{T}$ described in Section~\ref{binarytree} to generate the mathematical expressions and operator sequences,
the operator  $D_{ij}$ is a convolution operator,  $D_{ij} \tilde{u} = q_{ij} \otimes \tilde{u}$  with $\otimes$ being the convolution operation.    
Note that the operator  $D_{ij}u$ approximates   $\frac{\partial^{i+j} u}{\partial^i x \partial^j y}$.

\subsection{Burgers' equation with constant coefficients}\label{burgers-cc}

We consider the Burgers' equation with a periodic boundary condition defined on the domain $\Omega=[0,2\pi]^2$  used in   \cite{long2019pde}:  
$$
\begin{cases}\frac{\partial u}{\partial t} & =-u \frac{\partial u}{\partial x} -v \frac{\partial u}{\partial y}+\nu(\frac{\partial^2 u}{\partial x^2}+\frac{\partial^2 u}{\partial y^2}), \\ 
\frac{\partial v}{\partial t} & =-u \frac{\partial v}{\partial x} -v \frac{\partial v}{\partial y}+\nu(\frac{\partial^2 v}{\partial x^2}+\frac{\partial^2 v}{\partial y^2}), \\ 
\left.u\right|_{t=0} &=u_0(x, y),\\
\left.v\right|_{t=0} &=v_0(x, y),\end{cases}
$$
where $(t, x, y) \in [0,4] \times \Omega$ and $\nu=0.05$. The initial values $u_0(x, y)$ and $v_0(x, y)$ are random realizations of the following random function
$$
w(x, y)=\frac{2 w_0(x, y)}{\max _{x, y}\left|w_0(x, y)\right|}+c,
$$
where $w_0(x, y)=\sum_{|k|,|l| \leq 4} \lambda_{k, l} \cos (k x+l y)+\gamma_{k, l} \sin (k x+l y)$ with $\lambda_{k, l}$ and $\gamma_{k, l}$ as random variables following the standard normal distribution $\mathcal{N}(0, 1)$, and $c$ is a random variable following the uniform distribution $\mathcal{U}(-2, 2)$.

In our approach, the training data is generated using a finite difference scheme on a $256 \times 256$ mesh. We employ the second-order Runge-Kutta method with a time step size of $\delta t = \frac{1}{1600}$ for temporal discretization. For spatial discretization, we use the second-order upwind scheme for the gradient operator $\nabla$ and the central difference scheme for the Laplacian operator $\Delta$.

To generate noisy training data $\hat{u}$, we add noise to the true solution  as follows:
\begin{equation*}
\widehat{u}(t, x, y)=u(t, x, y)+C \times M \times W,
\end{equation*}
where $M = \max_{x, y, t}{u(t, x, y)}$, $W \sim \mathcal{N}(0, 1)$ is a random variable, and $C = 0.001$ represents the noise level.

Note that the order of the underlying PDE is no more than 2.  We use one $\operatorname{FEXTree}$ to learn the right-hand side of the Burgers' equation. We denote this $\operatorname{FEXTree}$ as $\mathrm{Tree}_u$. Each forward step of our $\operatorname{FEX}$ can be written as follows
$$
\begin{aligned}
& \tilde{u}\left(t_{i+1}, \cdot\right)=\tilde{u}\left(t_i, \cdot\right)+\delta t \cdot \operatorname{Tree}_u\left(D_{00} \tilde{u}, D_{01} \tilde{u}, \cdots, D_{20} \tilde{u}\right), \\ 
\end{aligned}
$$where $\{D_{ij}: 0 \leq i + j \leq 2\}$ represents the convolution operator capturing the differential operator   approximating $\frac{\partial^{i+j} u}{\partial^i x \partial^j y}$.

\begin{table}[htbp]
    \centering
    \caption{\textbf{Numerical Results of the Burger’s equation term by term}}
    \begin{tabular}{llllll}
        \toprule \text{True PDE} & \text{PDE-Net 2.0} & \text{SINDy} & \text{GP} & \text{SPL} & \text{FEX} \\
        \midrule $-uu_x$ & $-1.00uu_x$ & $0uu_x$ & $-1.00uu_x$ & $0uu_x$ & $-1.00uu_x$ \\
        \midrule $-vu_y$ & $-1.00vu_y$ & $0vu_y$ & $-1.00vu_y$ & $-1.00vu_y$ & $-1.00vu_y$ \\
        \midrule $0.05u_{xx}$ & $0.0503u_{xx}$ & $0.0832u_{xx}$ & $0u_{xx}$ & $0u_{xx}$ & $0.0498u_{xx}$\\
        \midrule $0.05u_{yy}$ & $0.0503u_{yy}$ & $0u_{yy}$ & $0u_{yy}$ & $0u_{yy}$ & $0.0502u_{yy}$\\
        \midrule & $5.98\times10^{-3}uv$ & $5.55\times10^{-1}u^2u_x$ & $7.6\times10^{-2}vu_{yy}$ & $u_x$ & $4.21\times10^{-4}u^2$\\
        \midrule & $1.56\times10^{-3}u$ & $-4.62\times10^{-1}u_yv^3$ & $-1.7\times10^{-2}v^2u_{yy}$ & 0 & $4\times10^{-4}u^2$\\
        \midrule & $-8.02\times10^{-4}uu_y$ & $-4.76e^{-1}u^2u_x$ & 0 & 0& $2\times10^{-4}u_x$\\
        \midrule & $-6.78e^{-4}uu_xu_y$ & $4.36\times10^{-2}uu_xu_{yy}$ & 0 & 0& $2\times10^{-4}u_{xy}$ \\
        \midrule & $-6.62\times10^{-4}v_x$ & $-4.26\times10^-2v^2v_y$ & 0 & 0& $2\times10^{-4}v_x$\\
        \midrule & $-5.82\times10^{-4}$ & $-3.5\times10^{-2}u_yvv_{yy}$ & 0 & 0& $2\times10^{-4}v_y$\\
        \bottomrule
    \end{tabular}
    \label{Burger_equa}
\end{table}

\begin{table}[htbp]
    \centering
    \caption{\textbf{Absolute errors for the Burger’s equation term by term}}
    \begin{tabular}{llllll}
        \toprule \text{True PDE} & \text{PDE-Net 2.0} & \text{SINDy} & \text{GP} & \text{SPL} & \text{FEX} \\
        \midrule $-uu_x$ & 0 & 1 & 0 & 1 & 0 \\
        \midrule $-vu_y$ & 0 & 1 & 0 & 0 & 0 \\
        \midrule $0.05u_{xx}$ & $3\times10^{-4}$ & $3.3\times10^{-3}$ & $5\times10^{-2}$ & $5\times10^{-2}$ & $2\times10^{-4}$\\
        \midrule $0.05u_{yy}$ & $3\times10^{-4}$ & $5\times10^{-2}$ & $5\times10^{-2}$ & $5\times10^{-2}$ & $2\times10^{-4}$\\
        \midrule & $5.98\times10^{-3}$ & $5.55\times10^{-1}$ & $7.6\times10^{-2}$ & 1 & $4.21\times10^{-4}$\\
        \midrule & $1.56\times10^{-3}$ & $4.62\times10^{-1}$ & $1.7\times10^{-2}$ & 0 & $4\times10^{-4}$\\
        \midrule & $8.02\times10^{-4}$ & $4.76\times10^{-1}$ & 0 & 0 & $2\times10^{-4}$\\
        \midrule & $6.78\times10^{-4}$ & $4.36\times10^{-2}$ & 0 & 0 & $2\times10^{-4}$ \\
        \midrule & $6.62\times10^{-4}$ & $4.26\times10^{-2}$ & 0 & 0 & $2\times10^{-4}$\\
        \midrule & $5.82\times10^{-4}$ & $3.5\times10^{-2}$ & 0 & 0 & $2\times10^{-4}$\\
        \bottomrule
    \end{tabular}
    \label{Burger error}
\end{table}

\begin{table}[htbp]
    \centering
    \caption{\textbf{Mean  absolute error for the Burger’s equation}}
    \begin{tabular}{llllll}
        \toprule & \text{PDE-Net 2.0} & \text{SINDy} & \text{GP} &\text{SPL} & \text{FEX}\\
        \midrule \text{Mean Absolute Error} & $1.086\times10^{-3}$ & $3.239\times10^{-1}$ & $1.93\times10^{-2}$ & $2.1\times10^{-1}$& $2.021\times10^{-4}$\\
        \bottomrule
    \end{tabular}
    \label{Burger mean error}
\end{table}


Table~\ref{Burger_equa} presents the target recovery terms of the Burgers' equation and the discovered terms by  PDE-Net $2.0$, SINDy, GP, SPL, and our FEX. Both our FEX and PDE-Net $2.0$ can accurately recover the governing equation and our FEX outperforms PDE-Net $2.0$ when the terms other than the dominant four terms are compared.  
In the case of SINDy, the hyperparameter $poly_{order}$ determines the highest degree of polynomials in its dictionary. The default value is $poly_{order}=5$ in SINDy, which leads to a large dictionary size of 6188, making it computationally expensive to run on an Apple M1 Pro CPU with 16GB of memory. In contrast, our FEX sets $poly_{order}=3$, resulting in a smaller dictionary size of 455 for candidate functions.
To provide results for GP, we utilize a genetic programming-based symbolic regression implementation available in the \texttt{gplearn} library.

 To evaluate the accuracy of these methods, the absolute error for each equation term and the mean absolute error (MAE) are shown in Table ~\ref{Burger error} and Table ~\ref{Burger mean error}, respectively. The absolute error for each term $w_i$ is computed as $|w_i-\tilde{w}_i|$, where $w_i$ is the true coefficient for the corresponding term in the true equation, and $\tilde{w}_i$ is the corresponding numerical coefficient obtained from each method. The MAE is the average of the absolute errors over all terms.
In this test, we consider a total of 10 terms, including the 4 dominating terms and 6 non-dominating terms.

We further assess the robustness of our FEX and PDE-Net 2.0 with different levels of noise. The results for these tests are provided in Table~\ref{Burger_equa_robust_FEX} for our FEX and in Table~\ref{Burger_equa_robust_PDENet} for PDE-Net 2.0. Our FEX is more robust and more accurate than PDE-Net 2.0 for higher levels of noise.

\begin{table}[htbp]
    \centering
    \caption{\textbf{Numerical Results of the Burger’s equation by FEX with different levels of noise}}
    \begin{tabular}{llll}
        \toprule \text{True  PDE} & $C=0.001$ & $C=0.005$ & $C=0.01$  \\
        \midrule $-uu_x$ & $-1.00uu_x$ & $-1.006uu_x$ & $-1.025uu_x$  \\
        \midrule $-vu_y$ & $-1.00vu_y$ & $-1.002vu_y$ & $-0.926vu_y$  \\
        \midrule $0.05u_{xx}$ & $0.0498u_{xx}$ & $0.0534u_{xx}$ & $0.0617u_{xx}$ \\
        \midrule $0.05u_{yy}$ & $0.0502u_{yy}$ & $0.0543u_{yy}$ & $0.0612u_{yy}$ \\
        \bottomrule
    \end{tabular}
    \label{Burger_equa_robust_FEX}
\end{table}

\begin{table}[htbp]
    \centering
    \caption{\textbf{Numerical Results of the Burger’s equation by PDE-Net with different levels of noise}}
    \begin{tabular}{llll}
        \toprule \text{True PDE} & $C=0.001$ & $C=0.005$ & $C=0.01$  \\
        \midrule $-uu_x$ & $-1.00uu_x$ & $-1.01uu_x$ & $-0.88uu_x$  \\
        \midrule $-vu_y$ & $-1.00vu_y$ & $-0.92vu_y$ & $-0.80vu_y$  \\
        \midrule $0.05u_{xx}$ & $0.0503u_{xx}$ & $0.01u_{xx}$ & $0.01u_{xx}$ \\
        \midrule $0.05u_{yy}$ & $0.0503u_{yy}$ & $0.02u_{yy}$ & $0.01u_{yy}$ \\
        \bottomrule
    \end{tabular}
    \label{Burger_equa_robust_PDENet}
\end{table}

\begin{figure}[htbp]
    \centerline{\includegraphics[width=\columnwidth]{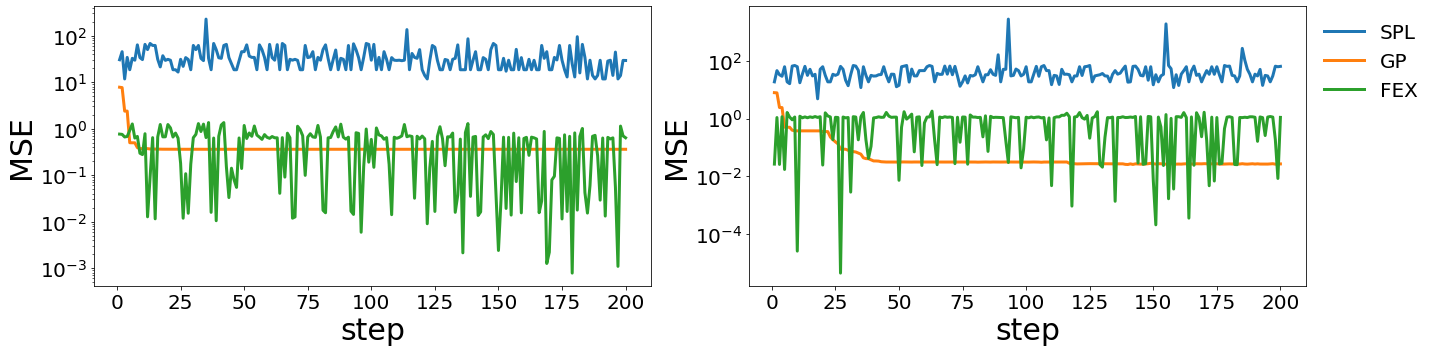}}
    \begin{center}
    \caption{Left: Worst results among 5 runs of three methods. Right: Best results among 5 runs of three methods}
    \label{lossdecay_burger}
    \end{center}
\end{figure}

Figure~\ref{lossdecay_burger} illustrates the training process of GP, SPL, and FEX for solving Burgers' equation. Table~\ref{lossdecay_burger_mean} presents the mean and best results obtained from 5 runs for each method.

\begin{table}[htbp]
    \centering
    \caption{\textbf{The best and average  Mean Squared Errors (MSE) for the Burger’s equation within 5 runs}}
    \begin{tabular}{llll}
        \toprule & \text{GP} &\text{SPL} & \text{FEX}\\
        \midrule \text{Best} & $0.026$ & $4.804$ & $4.575\times10^{-6}$\\
        \midrule \text{Average} & $0.159$ & $9.975$ & $2.134\times10^{-4}$\\
        \bottomrule
    \end{tabular}
    \label{lossdecay_burger_mean}
\end{table}

\subsection{Burgers' equation with varying coefficients }\label{burgers-vc}
We consider a typical example of the Burgers' equation with varying coefficients which seeks $u(x,t)$ such that
\begin{equation*}
    \begin{split}
    u_t (x,t)&=a(t) u u_x+\nu u_{x x}, \qquad \forall (x, t)\in[-8, 8] \times [0,10], \\
   u(x, 0)&=\exp (-(x+1)^2),
    \end{split}
\end{equation*} 
where $a(t) = 1 + \frac{1}{4} \sin(t)$ and $\nu = 0.1$. To solve this equation numerically, we use the \textit{odeint} function from the \texttt{SciPy} package and apply the discrete Fourier transform (DFT) to evaluate the spatial derivatives.

We add 1\% noise to generate data denoted by $\tilde{u}$. Suppose we know a priori of $\tilde{u}_x$ and $ \tilde{u}_{xx}$ and suppose the varying coefficients change with variables of $\tilde{u}$ (i.e. x and t). Our $\operatorname{FEX}$ can then be expressed as 
$$
\tilde{u}\left(t_{i+1}, \cdot\right)=\tilde{u}\left(t_i, \cdot\right)+\delta t \cdot \operatorname{Tree}_u(x, t, \tilde{u},\tilde{u}_x, \tilde{u}_{xx}),
$$
where $t_{i+1}=t_{i}+\delta t$, and $\operatorname{Tree}_u(x, t, \tilde{u}, \tilde{u}_x, \tilde{u}_{xx})$ denotes the FEX tree that approximates the right-hand side of the PDE based on the available data.

\begin{figure}[htbp]
    \centerline{\includegraphics[width=\columnwidth]{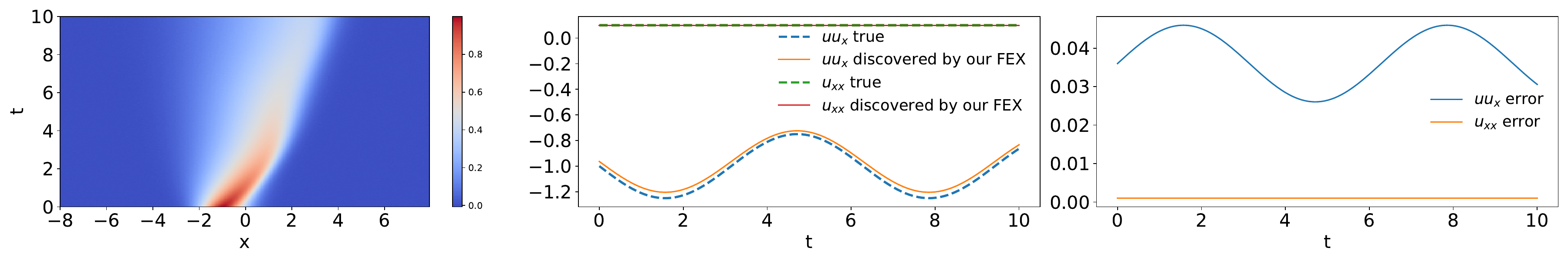}}
    \begin{center}
    \caption{Left: Our numerical solution of Burgers' equation with varying coefficients. Middle: Comparison of real and reconstructed coefficients of diffusion and advection term. Right: Absolute error.}
    \label{burger_vc}
    \end{center}
\end{figure}

The numerical results for Burgers' equation with varying coefficients, obtained from PDE-Net 2.0, SINDy, GP, SPL, and our FEX, are summarized in Table~\ref{burger_vc_result}. Table~\ref{burger_vc_result} shows that our FEX method can capture the varying coefficient in the nonlinear advection term, while other methods fail to detect this feature.
 Figure~\ref{burger_vc}   demonstrates that our FEX method can approximate both the constant coefficient and the time-varying coefficient in the PDE more accurately.
While recent works such as \cite{chen2021robust,luo2021ko,rudy2019data} have discovered the varying coefficients in PDEs, it is important to note that their numerical approximations are expressed in the numerical forms, while our FEX method presents the numerical solution in a symbolic/expression form.

\begin{table}[htbp]
    \begin{center}
    \caption{\textbf{Comparison for the Burgers' equation with varying coefficients.}}
    \label{burger_vc_result}
    \begin{tabular}{lll}
    \toprule \text {True PDE } & $u_t=-(1+\sin (t) / 4)uu_x+0.1u_{xx}$\\[1ex]
    \midrule
    \text{PDE-Net 2.0 } & $u_t = -0.780uu_x+0.104u_{xx}-0.326ux^3+0.316u^2u_x^2-0.225uu_x^2+\dots$\\[1ex]
    \midrule
    \text{SINDy } & $u_t=-0.948uu_x-0.100u_{xx}-0.168u^2u_x-0.087u_x^3+0.050uu_x^2-0.043u_x^2$\\[1ex]
    \midrule
    \text {GP } &$u_t = -1.011uu_x$\\[1ex]
    \midrule
    \text {SPL } &$u_t = -0.5491u_x$\\[1ex]
    \midrule
    \text {FEX } & $u_t=-0.964uu_x-0.249\sin (t)uu_x+0.099u_{xx}$\\
    \bottomrule
    \end{tabular}
    \end{center}
\end{table}

\begin{figure}[htbp]
    \centerline{\includegraphics[width=\columnwidth]{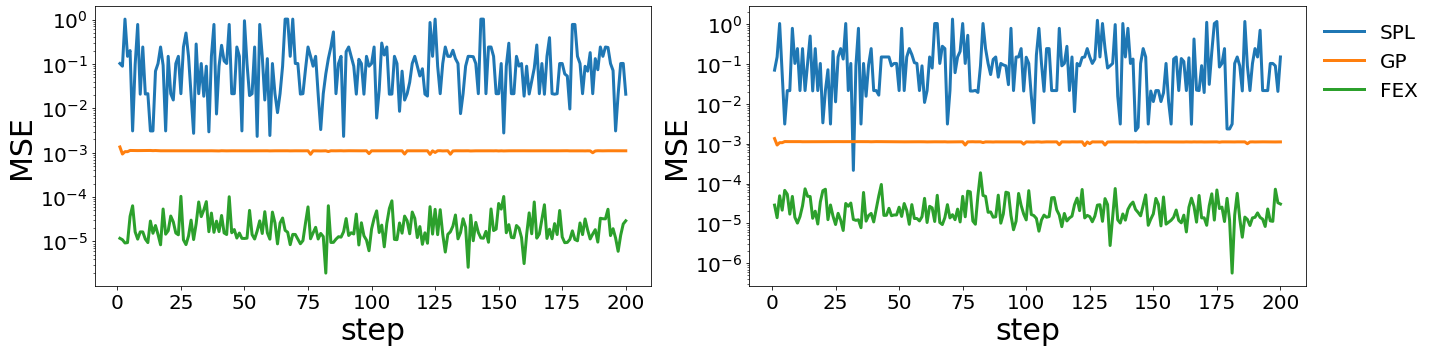}}
    \begin{center}
    \caption{Left: Worst results among 5 runs of three methods. Right: Best results among 5 runs of three methods}
    \label{lossdecay_burger_vc}
    \end{center}
\end{figure}

The training processes of GP, SPL, and FEX are demonstrated in Figure~\ref{lossdecay_burger_vc}.    Table~\ref{lossdecay_burger_vc_mean} shows the best and average  MSE among 5 runs for each method.

\begin{table}[htbp]
    \centering
    \caption{\textbf{The best and average MSE for the Burger’s equation with varying coefficients among 5 runs}}
    \begin{tabular}{llll}
        \toprule & \text{GP} &\text{SPL} & \text{FEX}\\
        \midrule \text{Best} & $8.940\times10^{-4}$ & $2.136\times10^{-4}$ & $5.698\times10^{-7}$\\
        \midrule \text{Average} & $9.044\times10^{-4}$ & $1.805\times10^{-3}$ & $1.338\times10^{-6}$\\
        \bottomrule
    \end{tabular}
    \label{lossdecay_burger_vc_mean}
\end{table}

\subsection{2D Hopf normal form}\label{hopf}

The 2D Hopf normal form is a classic example of a parameterized system that exhibits bifurcations as the parameter $\mu$ varies. The system is given as follows  \cite{brunton2016discovering} 
$$
\begin{cases}
    \begin{aligned}
& \dot{x}=\mu x+\omega y-A x\left(x^2+y^2\right), \\
& \dot{y}=-\omega x+\mu y-A y\left(x^2+y^2\right),
\end{aligned}
\end{cases}
$$
where $A=1$ and $\omega =1$. The parameter $\mu$ plays a crucial role in determining the system's behavior and affects whether the system exhibits stable or unstable dynamics. Experimental data is generated from unstable values of $\mu$ ($\mu = -0.15, -0.05$) and stable values of $\mu$ ($\mu = 0.05, 0.15, 0.25, 0.35, 0.45, 0.55$). These data samples will be used as training data for the FEX algorithm to discover the normal form for different  $\mu$.

\begin{table}[htbp]
    \begin{center}
    \caption{\textbf{Comparison for the 2D Hopf normal form.}}
    \label{Hopf form}
    \begin{tabular}{lll}
    \toprule \text {True equations } &$\dot{x}= \mu x- y-x\left(x^2+y^2\right)$ \\ & $\dot{y}=x+ \mu y-y\left(x^2+y^2\right)$\\[1ex]
    \midrule
     \text {PDE-Net 2.0 } & \makecell[l]{$\dot{x}=0.912\mu x-0.998y-0.930x^3-0.935xy^2$} \\& \makecell[l]{$\dot{y}=0.999x +0.960\mu y-0.979yx^2-0.930y^3$}\\
    \midrule
    \text{SINDy } & $\dot{x}= -0.993y$ \\& \makecell[l]{$\dot{y}= 0.995x+1.006\mu y-1.006yx^2-1.008y^3$}\\[1ex]
    \midrule
    \text{GP } & $\dot{x}= -y$ \\& \makecell[l]{$\dot{y}=x$}\\[1ex]
    \midrule
    \text{SPL } & $\dot{x}= -0.9391y$ \\& \makecell[l]{$\dot{y}=x$}\\[1ex]
    \midrule
    \text {FEX } & \makecell[l]{$\dot{x}=0.984\mu x-0.997y-1.003x^3-1.004xy^2$} \\& \makecell[l]{$\dot{y}=0.996x +1.014\mu y-0.995yx^2-1.000y^3$}\\
    \bottomrule
    \end{tabular}
    \end{center}
\end{table}

\begin{table}[htbp]
    \centering
    \caption{\textbf{Results of the 2D Hopf normal form term by term }}
    \resizebox{\columnwidth}{!}{%
    \begin{tabular}{llllll}
        \toprule \text{True PDE for $\dot{x}$ } & \text{PDE-Net 2.0} & \text{FEX} & \text{True PDE for $\dot{y}$ } & \text{PDE-Net 2.0} & \text{FEX}\\
        \midrule $\mu x$ & $0.912\mu x$ & $0.984\mu x$ & $x$ & $0.999x$ & $0.996x$\\
        \midrule $-y$ & $-0.998y$ & $-0.997y$ & $\mu y$ & $0.960\mu y$ & $1.014\mu y$ \\
        \midrule $-x^3$ & $-0.930x^3$ & $-1.003x^3$ & $-yx^2$ & $-0.979yx^2$ & $-0.995yx^2$\\
        \midrule $-xy^2$ & $-0.935xy^2$ & $-1.004xy^2$ & $-y^3$ & $-0.930y^3$ & $-1.000y^3$\\
        \midrule & $-3.33\times 10^{-1}xy^3$ & $7.66\times 10^{-2}y^3$ && $-2.45\times 10^{-1}\mu y^3$ & $1.75\times 10^{-2}\mu^3$\\
        \midrule & $3.22\times 10^{-1}\mu xy$ & $6.35\times 10^{-2}\mu y$ && $1.25\times 10^{-1}x^2y^2$ & $-4.05\times10^{-3}\mu^2$ \\
        \midrule & $-3.20\times 10^{-1}x^3y$ & $2.16\times 10^{-2}x\mu^2$ && $1.20\times 10^{-1}y^4$ & $3.99\times10^{-3}\mu x^2$ \\
        \midrule & $3.09\times 10^{-1}y^4$ & $-5.41\times10^{-3}\mu^3$ && $1.13\times 10^{-1}\mu xy$ & $2.80\times10^{-3}xy$\\
        \midrule & $-3.04\times 10^{-1}\mu y^2$ & $2.70\times10^{-3}x$ && $-1.09\times 10^{-1}\mu xy^2$ & $2.20\times10^{-3}y$\\
        \midrule & $2.72\times 10^{-1}x^2y^2$ & $2.16\times10^{-3}x\mu^2$ && $-1.03\times 10^{-1}xy^4$ & $1.64\times10^{-3}x^3$\\
        \bottomrule
    \end{tabular}
    }
    \label{Hopf_per_parameter}
\end{table}

\begin{table}[htbp]
    \centering
    \caption{\textbf{Absolute error of the 2D Hopf normal form term by term }}
    \resizebox{\columnwidth}{!}{%
    \begin{tabular}{llllll}
        \toprule \text{True PDE  for $\dot{x}$ } & \text{PDE-Net 2.0  } & \text{FEX  } & \text{True PDE for  $\dot{y}$ } & \text{PDE-Net 2.0  } & \text{FEX }\\
        \midrule $\mu x$ & $8.80\times10^{-2}$ & $1.60\times10^{-2}$ & $x$ & $1.00\times10^{-3}$ & $4.00\times10^{-3}$\\
        \midrule $-y$ & $2.00\times10^{-3}$ & $3.00\times10^{-3}$ & $\mu y$ & $4.00\times10^{-2}$ & $1.40\times10^{-2}$ \\
        \midrule $-x^3$ & $7.00\times10^{-2}$ & $3.00\times10^{-3}$ & $-yx^2$ & $2.10\times10^{-2}$ & $5.00\times10^{-3}$\\
        \midrule $-xy^2$ & $6.50\times10^{-2}$ & $4.00\times10^{-3}$ & $-y^3$ & $7.00\times10^{-2}$ & $0$\\
        \midrule & $3.33\times 10^{-1}xy^3$ & $7.66\times 10^{-2}y^3$ && $2.45\times 10^{-1}\mu y^3$ & $1.75\times 10^{-2}\mu^3$\\
        \midrule & $3.22\times 10^{-1}\mu xy$ & $6.35\times 10^{-2}\mu y$ && $1.25\times 10^{-1}x^2y^2$ & $4.05\times10^{-3}\mu^2$ \\
        \midrule & $3.20\times 10^{-1}x^3y$ & $2.16\times 10^{-2}x\mu^2$ && $1.20\times 10^{-1}y^4$ & $3.99\times10^{-3}\mu x^2$ \\
        \midrule & $3.09\times 10^{-1}y^4$ & $5.41\times10^{-3}\mu^3$ && $1.13\times 10^{-1}\mu xy$ & $2.80\times10^{-3}xy$\\
        \midrule & $3.04\times 10^{-1}\mu y^2$ & $2.70\times10^{-3}x$ && $1.09\times 10^{-1}\mu xy^2$ & $2.20\times10^{-3}y$\\
        \midrule & $2.72\times 10^{-1}x^2y^2$ & $2.16\times10^{-3}x\mu^2$ && $1.03\times 10^{-1}xy^4$ & $1.64\times10^{-3}x^3$\\
        \bottomrule
    \end{tabular}
    }
    \label{Hopf_per_parameter_error}
\end{table}

\begin{table}[htbp]
    \centering
    \caption{\textbf{Mean absolute error of  the 2D Hopf normal form}}
    \begin{tabular}{lllll}
        \toprule & \text{PDE-Net} for $\dot{x}$    & \text{FEX} for $\dot{x}$  & \text{PDE-Net} for $\dot{y}$  & \text{FEX} for $\dot{x}$ \\
        \midrule \text{Mean Absolute Error} & $2.085\times 10^{-1}$ & $1.977\times10^{-2}$ & $9.470\times10^{-2}$ & $5.518\times10^{-3}$\\
        \bottomrule
    \end{tabular}
    \label{Hopf mean error}
\end{table}

\begin{figure}[htbp]
\centerline{\includegraphics[width=0.8\columnwidth]{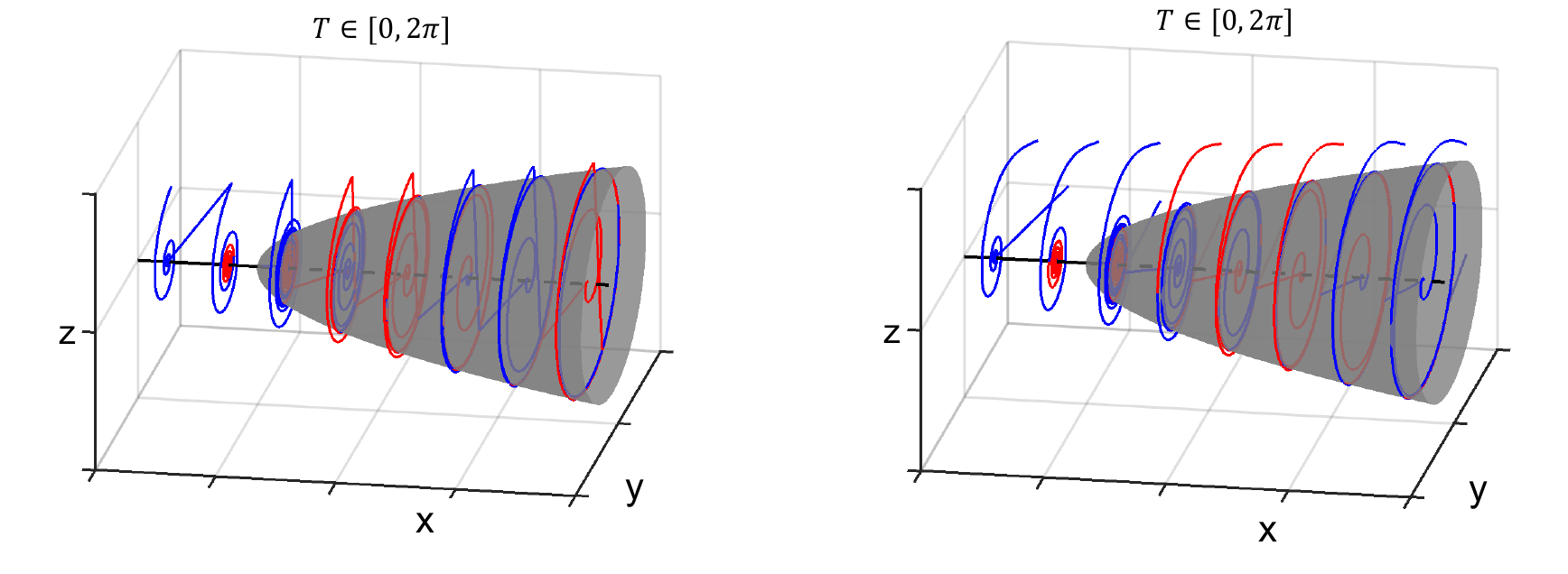}}
    \begin{center}
    \caption{The visualization of the ground truth dynamics of the 2D Hopf normal form (left) and its prediction learned by FEX (right).}
    \label{hopf pred}
    \end{center}
\end{figure}

 The results obtained from the FEX algorithm for the 2D Hopf normal form demonstrate its capability to accurately recover the analytical form of the unknown dynamical system model for different values of the bifurcation parameter $\mu$.  
We can see from Table~\ref{Hopf form}  that both PDE-Net 2.0 and our FEX can recover the desired governing equation, while SINDy, GP, and SPL fail.  Table~\ref{Hopf_per_parameter} demonstrates four dominating terms and six additional terms in descending coefficients for $\dot{x}$ and $\dot{y}$  generated by PDE-Net 2.0 and FEX. It is observed that FEX outperforms PDE-Net 2.0 in terms of accuracy, as the coefficients of the six additional terms generated by our FEX are smaller than those generated by PDE-Net 2.0.
Tables~\ref{Hopf_per_parameter_error} and \ref{Hopf mean error}  present the absolute error term by term and the mean absolute error for PDE-Net 2.0 and FEX. We can see from Tables~\ref{Hopf_per_parameter_error} and \ref{Hopf mean error} that  FEX achieves a smaller absolute error term by term and a smaller MAE than that of  PDE-Net 2.0.  Figure~\ref{hopf pred} showcases the dynamics predicted by FEX and the ground truth dynamics when $T\in[0,2\pi]$. We can see from  Figure~\ref{hopf pred} that the governing equation learned by our FEX performs well in predicting and capturing the underlying dynamics of the system.

\begin{figure}[htbp]
    \centerline{\includegraphics[width=\columnwidth]{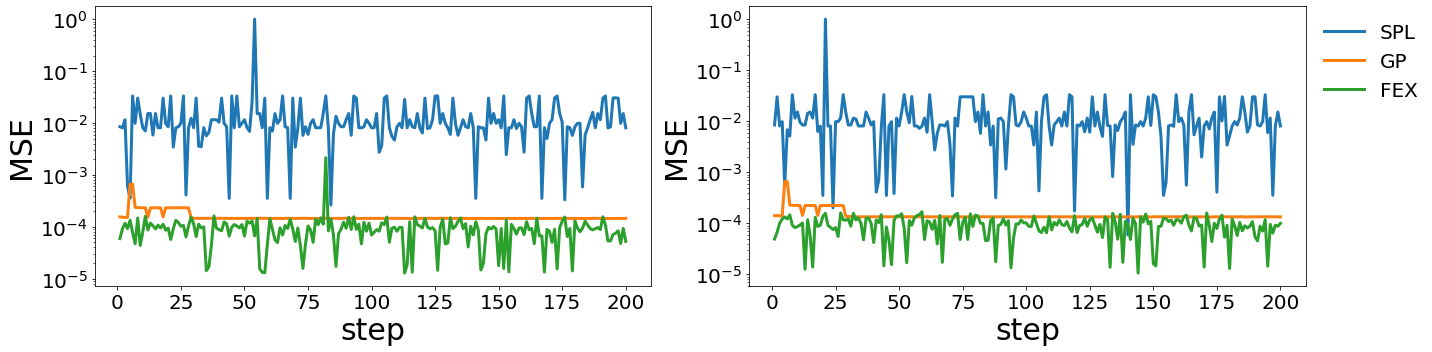}}
    \begin{center}
    \caption{Left: Worst results among 5 runs of three methods. Right: Best results among 5 runs of three methods}
    \label{lossdecay_hopf_x}
    \end{center}
\end{figure}

Figure~\ref{lossdecay_hopf_x} demonstrates the training processes of GP, SPL, and our FEX for the 2D Hopf normal form. We can see from Figure~\ref{lossdecay_hopf_x}   that the average squared error of our FEX is the smallest compared with that of GP and SPL. 
 Table~\ref{lossdecay_hopf_x_mean} shows the best and average MSE obtained from 5 runs for each method, where our FEX provides the smallest best and average  MSE compared with those of GP and SPL.    

\begin{table}[htbp]
    \centering
    \caption{\textbf{The best and average  MSE for the 2D Hopf normal form  among 5 runs}}
    \begin{tabular}{llll}
        \toprule & \text{GP} &\text{SPL} & \text{FEX}\\
        \midrule \text{Best} & $1.329\times10^{-4}$ & $5.916\times10^{-5}$ & $1.058\times10^{-5}$\\
        \midrule \text{Average} & $1.386\times10^{-4}$ & $1.207\times10^{-4}$ & $1.169\times10^{-5}$\\
        \bottomrule
    \end{tabular}
    \label{lossdecay_hopf_x_mean}
\end{table}

\subsection{Johnson-Mehl-Avrami-Kolmogorov (JMAK) nonlinear  equation}\label{Avrami}
 JMAK plays a crucial role in describing the growth kinetic of phases at a constant temperature in material science. The Avrami equation, a specific form of  JMAK, is given by:
$$
y=1-\exp \left(-k t^n\right),
$$ 
where the coefficients $k$ and $n$ vary depending on the specific environmental conditions.

 In our experiment, we shall simulate the Avrami equation in different environmental conditions. We take $n=2$ and treat $k$ as a variable in the function $y = f(k, t)$. To create a dataset for training, we select multiple values  $k =$ {0.005, 0.01, 0.04, 0.1, 0.5, 0.8} and generate 20 uniformly distributed samples in the range $t \in [0, 1]$ for each $k$.

Using our $\operatorname{FEX}$ algorithm, we learn the Avrami equation using the generated dataset which is denoted by 
$$
\tilde{y}=\operatorname{Tree}_y(t,k),
$$
where $\tilde{y}$ denotes the predicted output based on the input variables $t$ and $k$.

\begin{table}[htbp]
    \begin{center}
    \caption{\textbf{Comparison for the Avrami equation with different $k$.}}
    \label{JMAK_k}
    \begin{tabular}{lll}
    \toprule \text {True function } &$y=1-\exp \left(-kt^2\right)$\\[1ex]
    \midrule
    \text{PDE-Net 2.0 } & $y=0.2613tk+0.7284t^2k-0.2751t^2k^2\cdots$\\[1ex]
    \midrule
    \text{SINDy } & $y=-0.0285tk-0.0435tk+1.0385t^2k+\cdots$\\[1ex]
    \midrule
    \text{GP } & $y=0.829kt^2$\\[1ex]
    \midrule
    \text{SPL } & $y=0.789kt^2$\\[1ex]
    \midrule
    \text {FEX } & $y=0.9996-\exp \left(-0.9997k t^2\right)$\\
    \bottomrule
    \end{tabular}
    \end{center}
\end{table}

\begin{figure}[htbp]
    \centerline{\includegraphics[width=\columnwidth]{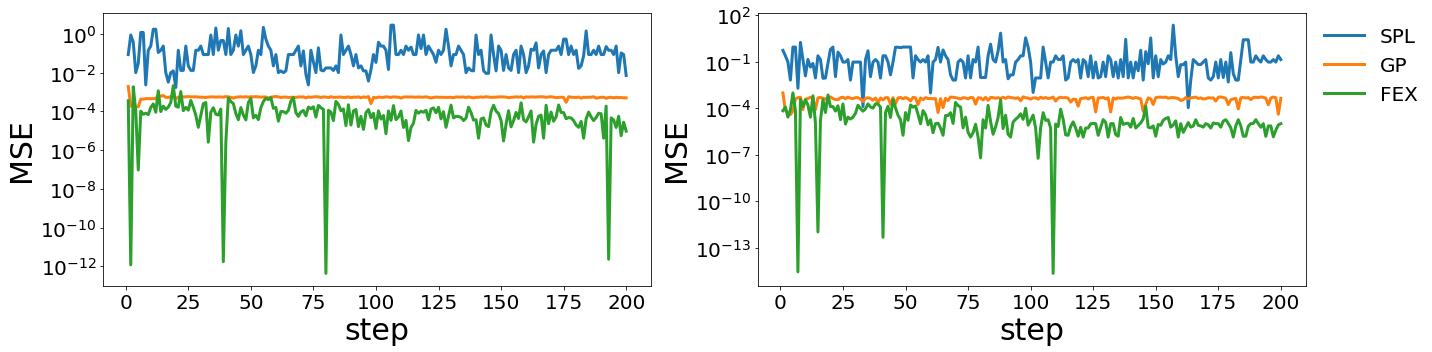}}
    \begin{center}
    \caption{Left: Worst results among 5 runs of three methods. Right: Best results among 5 runs of three methods}
    \label{lossdecay_avrami_k}
    \end{center}
\end{figure}

 We can see from Table~\ref{JMAK_k} that the equation learned by our FEX algorithm is the most accurate; while PDE-Net, GP, SINDy, and SPL fail to discover the equation.
Figure~\ref{lossdecay_avrami_k}  illustrates the worst and best results among 5 runs when GP, SPL, and FEX  are employed to discover the Avrami equation with varying coefficient $k$. We can see from Figure~\ref{lossdecay_avrami_k}  that our FEX has the smallest squared error compared to that of GP and SPL. Additionally,   We can see from Table~\ref{lossdecay_avrami_k_mean}  that our FEX offers the smallest best and average MSE  results compared to those of GP and SPL.

\begin{table}[htbp]
    \centering
    \caption{\textbf{The best and average  MSE for the Avrami equation with varying coefficients among 5 runs}}
    \begin{tabular}{llll}
        \toprule & \text{GP} &\text{SPL} & \text{FEX}\\
        \midrule \text{Best} & $3.936\times10^{-5}$ & $1.103\times10^{-4}$ & $2.169\times10^{-15}$\\
        \midrule \text{Average} & $7.184\times10^{-5}$ & $5.329\times10^{-4}$ & $1.169\times10^{-13}$\\
        \bottomrule
    \end{tabular}
    \label{lossdecay_avrami_k_mean}
\end{table}

In summary, PDE-Net 2.0 is capable of handling Burgers' equation
with constant coefficients and 2D Hopf normal form. SINDY can only discover the equation for $\dot{y}$ in 2D Hopf normal form.  GP and SPL methods show poor performance in all four cases. However, our FEX can discover Burgers' equation with constant coefficients and varying coefficients as well as handle the dynamic system of 2D Hopf normal form and non-linear JMAK equation. A comparison is presented in Table~\ref{results table}.

\begin{table}[htbp]
    \centering
    \caption{\textbf{Experimental results of different models discovering governing equations from data}}
    \resizebox{\textwidth}{!}{\begin{tabular}{lllllll}
        \toprule & \text{Correct equation structure} & \text{PDE-Net 2.0} & \text{SINDy} & \text{GP} & \text{SPL} & \text{FEX} \\
        \midrule \makecell[l]{Burger’s equation \\with constant coefficients} &
         $\begin{cases}
            \frac{\partial u}{\partial t} & =-u \frac{\partial u}{\partial x} -v \frac{\partial u}{\partial y}+0.05(\frac{\partial^2 u}{\partial x^2}+\frac{\partial^2 u}{\partial y^2}), \\ 
            \frac{\partial v}{\partial t} & =-u \frac{\partial v}{\partial x} -v \frac{\partial v}{\partial y}+0.05(\frac{\partial^2 v}{\partial x^2}+\frac{\partial^2 v}{\partial y^2}),\end{cases}$ 
         & $\sqrt{ }$ &  &  &  & $\sqrt{ }$ \\
        \midrule \makecell[l]{Burger’s equation \\with varying coefficients} & $u_t (x,t)=(1 + \frac{1}{4} \sin(t)) u u_x+0.1u_{x x}$ & & & & & $\sqrt{ }$ \\
        \midrule \text{2D Hopf normal form} & 
        $\begin{cases}
            \begin{aligned}
            & \dot{x}=\mu x+y-x\left(x^2+y^2\right), \\
            & \dot{y}=- x+\mu y-y\left(x^2+y^2\right),
            \end{aligned}
            \end{cases}$
        & $\sqrt{ }$ & $\sqrt{}\mkern-9mu{\smallsetminus}$ & & & $\sqrt{ }$\\
        \midrule \text{JMAK equation} & $y=1-\exp \left(-k t^2\right)$ & & & & & $\sqrt{ }$\\
        \bottomrule
    \end{tabular}}
    \label{results table}
\end{table}
\section{Conclusion} In this paper, we introduce a novel approach called the "finite expression method" (FEX)  to discover governing equations from data. Different from traditional black-box deep learning methods, FEX is a  symbolic learning method and provides interpretable and physically meaningful formulas.
The key strength of FEX lies in its ability to automatically generate a diverse range of mathematical expressions using a compact set of predefined operators. This flexibility allows FEX to capture underlying patterns and relationships in the data accurately, leading to precise equation discovery.
To validate the effectiveness of FEX, we conducted extensive numerical tests, comparing it with existing symbolic approaches such as PDE-Net, SINDy, SPL, and GP. The results demonstrate FEX's superiority in terms of accuracy and equation recovery. 
In conclusion, our proposed FEX approach represents a significant advancement in the field of symbolic learning, offering an effective and interpretable solution for equation discovery. The high accuracy and wide applicability of our FEX make it a valuable tool in diverse scientific and engineering domains.

 \section*{Acknowledgements}
C. W. was partially supported by National Science Foundation under awards DMS-2136380 and DMS-2206332. H. Y. was partially supported by the US National Science Foundation under awards DMS-2244988, DMS-2206333, the Office of Naval Research Award N00014-23-1-2007, and the DARPA D24AP00325-00. We would like to express our gratitude to ChatGPT for enhancing the wording during the paper-writing phase. Approved for public release; distribution is unlimited.

\bibliographystyle{unsrt}  
\bibliography{references}  

\section{Appendix}
\subsection{Trainable convolution filters}\label{trainable convolution filters}

In this context, the moment matrix for a $N \times N$ convolution filter $q$ is defined as follows:
$$
M(q)=\left(m_{i, j}\right)_{N \times N},
$$
where
$$
m_{i, j}=\frac{1}{i ! j !} \sum_{k_1, k_2=-\frac{N-1}{2}}^{\frac{N-1}{2}} k_1^i k_2^j q\left[k_1, k_2\right], i, j=0,1, \ldots, N-1.
$$
When applying the convolution filter $q$ to a smooth function $f: \mathbb{R}^2 \rightarrow \mathbb{R}$, we can use Taylor's expansion to obtain the following formula (as shown in \cite{long2019pde}):
\begin{equation*}
\begin{aligned}
& \sum_{k_1, k_2=-\frac{N-1}{2}}^{\frac{N-1}{2}} q\left[k_1, k_2\right] f\left(x+k_1 \delta x, y+k_2 \delta y\right) \\
= & \left.\sum_{k_1, k_2=-\frac{N-1}{2}}^{\frac{N-1}{2}} q\left[k_1, k_2\right] \sum_{i, j=0}^{N-1} \frac{\partial^{i+j} f}{\partial^i x \partial^j y}\right|_{(x, y)} \frac{k_1^i k_2^j}{i ! j !} \delta x^i \delta y^j+o\left(|\delta x|^{N-1}+|\delta y|^{N-1}\right) \\
= & \left.\sum_{i, j=0}^{N-1} m_{i, j} \delta x^i \delta y^j \cdot \frac{\partial^{i+j} f}{\partial^i x \partial^j y}\right|_{(x, y)}+o\left(|\delta x|^{N-1}+|\delta y|^{N-1}\right).
\end{aligned}
\end{equation*}
This expression allows us to design the convolution filter $q$ to approximate any differential operator with a prescribed order of accuracy by imposing constraints on its moment matrix $M(q)$.

For example, if we aim to design a $5 \times 5$ convolution filter $q$ to approximate the discretization of $\frac{\partial^2 }{\partial y^2}$, we can impose specific constraints on its moment matrix $M(q)$. The following two moment matrices illustrate different levels of approximation accuracy:
$$
\begin{pmatrix} 0 &0 &1 &0 &\star \\0 &0 &0 &\star &\star   \\0 &0  &\star &\star &\star \\ 0 &\star &\star &\star &\star  \\ \star &\star &\star &\star &\star \end{pmatrix}\qquad or \qquad \begin{pmatrix} 0 &0 &1 &\star &\star \\0 &0 &\star &\star &\star   \\0 &\star  &\star &\star & \star \\ \star &\star &\star &\star &\star  \\ \star &\star &\star &\star &\star \end{pmatrix}.
$$
In the above matrices, $\star$ denotes an unconstrained entry. The constraints imposed on the moment matrix on the left guarantee an approximation accuracy of at least second order, while the constraints on the moment matrix on the right ensure an approximation accuracy of at least first order.

\end{document}